\PassOptionsToPackage{table}{xcolor}
\documentclass{article} % For LaTeX2e
\usepackage{iclr2027_conference,times}

\usepackage{amsmath,amsfonts,bm}

\def\eqref#1{equation~\ref{#1}}
\def\1{\bm{1}}

\DeclareMathAlphabet{\mathsfit}{\encodingdefault}{\sfdefault}{m}{sl}
\SetMathAlphabet{\mathsfit}{bold}{\encodingdefault}{\sfdefault}{bx}{n}

\usepackage[utf8]{inputenc} % allow utf-8 input
\usepackage[T1]{fontenc}    % use 8-bit T1 fonts
\usepackage{wrapfig}
\usepackage{url}            % simple URL typesetting
\usepackage{booktabs}       % professional-quality tables
\usepackage{amsfonts}       % blackboard math symbols
\usepackage{nicefrac}       % compact symbols for 1/2, etc.
\usepackage{microtype}      % microtypography
\usepackage{xcolor}         % colors and table cell shading
\usepackage{graphicx}
\usepackage{subfigure}
\usepackage{mathtools}
\usepackage{amsthm}
\usepackage{amsmath}
\usepackage{amssymb}
\usepackage{diagbox}
\usepackage{multirow}
\usepackage[most]{tcolorbox}
\usepackage{makecell}
\definecolor{text}{rgb}{0.4,0.1,0.4}

\usepackage{float}
\usepackage{capt-of}
\usepackage{array}
\usepackage{tabularx}
\usepackage{placeins}
\usepackage{xurl} % Allow long bibliography URLs to wrap.
\usepackage{hyperref}       % hyperlinks
\hypersetup{
  colorlinks=true,
  linkcolor=NUSBlue,
  citecolor=NUSBlue,
  urlcolor=NUSBlue
}

\newcommand{\method}{\textsc{LPA-CWM}}

\title{
\method: A Learned Physical Adjudicator\\[-0.12em]
for Motion Reasoning with Counterfactual World Models
}

\author{%
\begin{tabularx}{\textwidth}{@{}*{4}{>{\centering\arraybackslash}X}@{}}
\authname{Kunwei Wu\textsuperscript{1,*}} &
\authname{Xiang Liu\textsuperscript{2,*}} &
\authname{Guocai Yao\textsuperscript{3}} &
\authname{Junming Chen\textsuperscript{4}} \\[0.10em]
\authname{Zhikang Chen\textsuperscript{5}} &
\authname{Min Zhang\textsuperscript{6}} &
\authname{Pengwei Wang\textsuperscript{3}} &
\authname{Sen Cui\textsuperscript{2,\(\dagger\)}}
\end{tabularx}
\\[0.30cm]
\begin{tabular}{@{}>{\centering\arraybackslash}p{0.45\textwidth}@{\hspace{0.01\textwidth}}>{\centering\arraybackslash}p{0.19\textwidth}@{\hspace{0.01\textwidth}}>{\centering\arraybackslash}p{0.32\textwidth}@{}}
\affilfont{\textsuperscript{1}National University of Singapore} &
\affilfont{\textsuperscript{2}Tsinghua University} &
\affilfont{\textsuperscript{3}Beijing Academy of Artificial Intelligence} \\
\affilfont{\textsuperscript{4}The Hong Kong University of Science and Technology} &
\affilfont{\textsuperscript{5}University of Oxford} &
\affilfont{\textsuperscript{6}East China Normal University}
\end{tabular}
\\[0.24cm]
{\footnotesize
\textcolor{NUSBlue}{\textbf{\textsuperscript{*}Equal contribution}}
\qquad
\textcolor{NUSBlue}{\textbf{\textsuperscript{\(\dagger\)}Project Leader \& Corresponding Author}}
}
\\[0.10cm]
{\footnotesize\sffamily
\textbf{Contact:}\enspace
\href{mailto:wukunwei@u.nus.edu}{\authmail{wukunwei@u.nus.edu}}\hspace{0.35em}\textbar\hspace{0.35em}
\href{mailto:xiang-liu25@mails.tsinghua.edu.cn}{\authmail{xiang-liu25@mails.tsinghua.edu.cn}}\hspace{0.35em}\textbar\hspace{0.35em}
\href{mailto:cuis@mail.tsinghua.edu.cn}{\authmail{cuis@mail.tsinghua.edu.cn}}
}
}

\begin{document}

\raggedbottom
\maketitle

\definecolor{bestbg}{HTML}{D9EAD3}
\definecolor{secondbg}{HTML}{FFF2CC}
\definecolor{thirdbg}{HTML}{F4CCCC}

\newcommand{\bestresult}[1]{\cellcolor{bestbg}\textbf{\strut #1}}
\newcommand{\secondresult}[1]{\cellcolor{secondbg}\strut #1}
\newcommand{\thirdresult}[1]{\cellcolor{thirdbg}\strut #1}
\newcommand{\tablecaptionspace}{\vspace{0.12cm}}
\newcommand{\tablebottomspace}{\vspace{-0.12cm}}

\begin{abstract}
Counterfactual world models (CWM) extract motion from pretrained video predictors by comparing factual and intervened predictions, but uniform aggregation weights responses equally without explicitly incorporating physical priors. Our key insight is to incorporate physical priors into candidate reliability learning, motivating \textbf{LPA-CWM} with a lightweight Learned Physical Adjudicator (LPA). Trained on dense MOVi-F trajectories, the 3.0M-parameter LPA compares visual context and response structure across an unordered candidate set to predict relative weights; windowed localization and one paired re-evaluation recover motion with the CWM frozen. Existing video-level benchmarks do not directly assess motion correspondence, where low localization error can conceal missing trajectory segments. We introduce Completeness-aware Motion Correspondence (CMC), a ground-truth-anchored protocol jointly measuring localization, completeness, visibility, and continuity, counting missing predictions as failures on visible dynamic points. Across DAVIS, Kinetics, and RoboTAP, LPA-CWM improves all main CMC measures over Uniform CWM, with relative gains of 18.1\%--60.0\% in average Dynamic Correspondence Accuracy ($\mathrm{DCA}_{\mathrm{avg}}$), and improves TAP-Vid First tracking accuracy (overview: \url{https://LPA-CWM.github.io}).
\end{abstract}

% Keep Figure 1 below the abstract on page 1; the introduction starts on page 2.
\begingroup
\setlength{\intextsep}{6pt}
\vspace{-0.35cm}
\begin{figure}[H]
\centering
\includegraphics[width=0.86\linewidth]{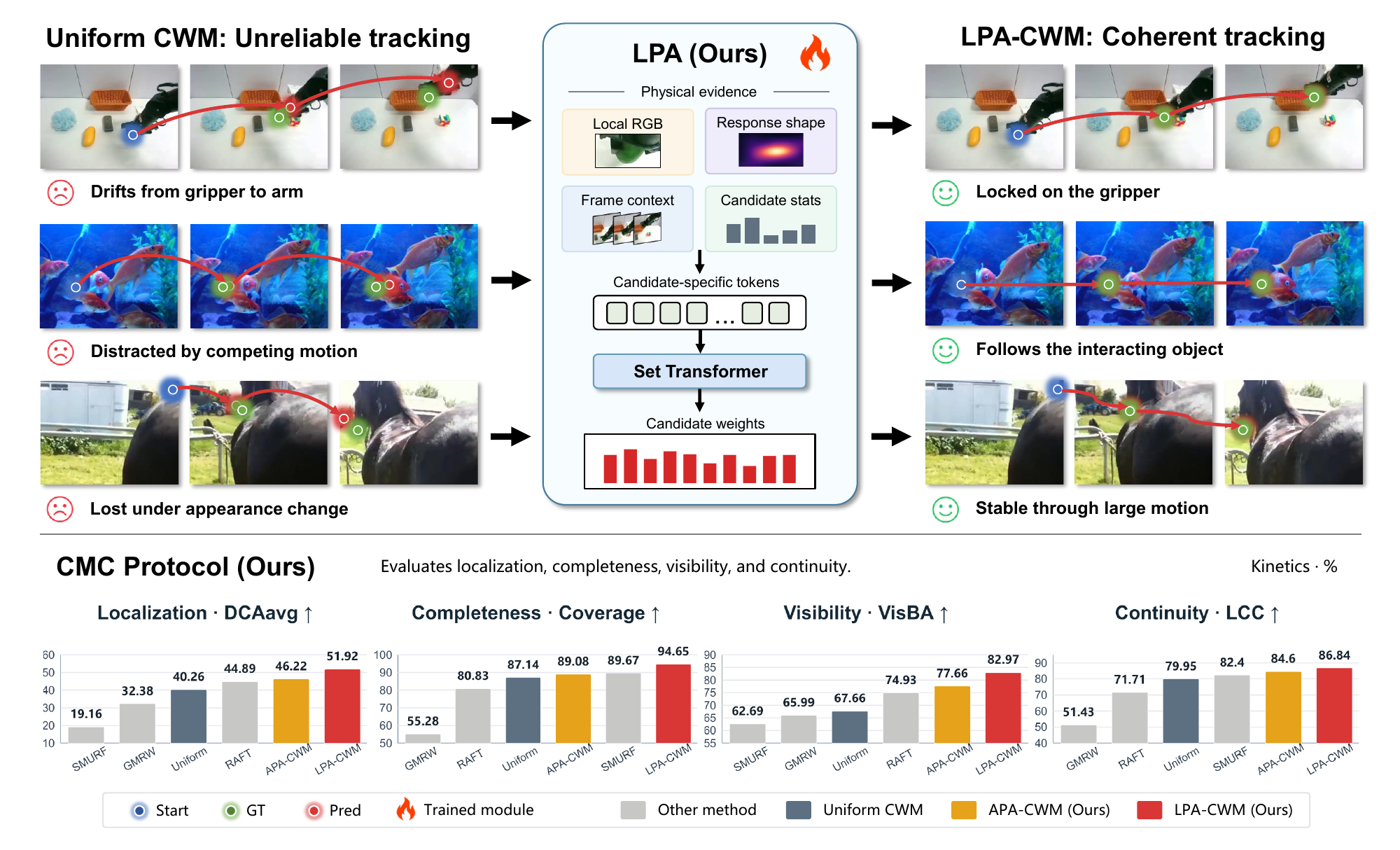}
\vspace{-0.55cm}
\caption{\footnotesize\textbf{Overview of our work.}
Uniform CWM can suffer from three characteristic failures in motion reasoning: drifting from moving objects, being distracted by competing motion, and losing track under appearance changes. Our Learned Physical Adjudicator (LPA) identifies reliable counterfactual motion evidence, enabling LPA-CWM to recover more coherent trajectories. We further introduce the CMC protocol to evaluate motion correspondence across localization, completeness, visibility, and continuity. Compared with external methods, Uniform CWM, and APA-CWM, LPA-CWM performs strongly across all four dimensions.}
\label{fig:teaser}
\vspace{-0.75cm}
\end{figure}
\endgroup
\clearpage

\section{Introduction}

World models learn predictive representations of visual dynamics~\citep{ha2018worldmodels,hafner2019planet,hafner2020dreamer}. Counterfactual World Models (CWM) extract motion from these representations by comparing predictions with and without localized interventions~\citep{bear2023cwm,venkatesh2023physical,kim2025motion}. Different target-frame masks can produce inconsistent responses for the same query: some localize the motion well, while others are affected by background content, competing motion, or visual ambiguity. Uniform aggregation treats all responses as equally reliable and can dilute useful evidence. This limitation partly stems from insufficient use of physical priors, motivating physically informed aggregation and comprehensive trajectory evaluation.

To incorporate physical priors into CWM inference, we formulate multi-mask aggregation as candidate reliability estimation. We first derive APA-CWM, an \textbf{Analytic Physical Adjudicator (APA)}, from the Normalized Spatiotemporal Gradient (NSG)~\citep{zhang2025nsg}. APA reads the NSG-derived spatiotemporal consistency around each candidate response peak and converts these local scores into closed-form reliability weights. It provides a training-free physical reference for testing how far an analytic consistency prior can resolve candidate ambiguity. We then propose LPA-CWM, whose \textbf{Learned Physical Adjudicator (LPA)} combines local visual evidence, response morphology, shared frame-pair context, and candidate statistics. A Transformer without candidate-index positional embeddings jointly compares the unordered candidate set and predicts relative weights.

We train only the 3.0M-parameter LPA on dense Kubric MOVi-F trajectories~\citep{greff2022kubric} to learn physical motion priors. For each query, dense forward motion provides the oracle target endpoint; candidate endpoint errors define relative reliability targets. Training combines distribution matching, weighted-coordinate localization, and best-candidate ranking. This gives direct supervision for which CWM responses are trustworthy while leaving the pretrained CWM unchanged. At inference, LPA weights the initial responses, followed by windowed localization and one paired CWM re-evaluation to obtain the final motion.

Existing benchmarks assess video quality, temporal consistency, physical and commonsense faithfulness, controllability, embodied utility, and counterfactual consistency~\citep{huang2024vbench,zheng2025vbench2,duan2025worldscore,shang2026worldarena,begiristain2026cronos}. They broaden evaluation beyond visual plausibility, but do not directly measure point correspondence recovered through CWM interventions. In motion recovery, low error on returned predictions can hide missing trajectory segments, while smooth trajectories may still follow incorrect motion.

We therefore introduce \textbf{Completeness-aware Motion Correspondence (CMC)}, a ground-truth-anchored protocol for annotated dynamic point tracks. CMC evaluates localization, completeness, visibility, and temporal continuity. Dynamic Correspondence Accuracy requires visible predictions within a distance threshold and counts missing predictions as failures on ground-truth-visible points. Track-level success further requires sufficient coverage and bounded localization error, while visibility and continuity diagnostics expose interrupted trajectories. CMC complements TAP-Vid First~\citep{doersch2022tapvid} by explicitly measuring completeness on moving tracks.

We evaluate on DAVIS, Kinetics, and RoboTAP~\citep{ponttuset2017davis,kay2017kinetics,vecerik2024robotap} against six flow, correspondence, and point-tracking baselines~\citep{stone2021smurf,shrivastava2024gmrw,jiang2023doduo,karaev2025cotracker3,teed2020raft,wang2024sea}. Across the three datasets, LPA-CWM improves all main CMC measures over Uniform CWM, with relative $\mathrm{DCA}_{\mathrm{avg}}$ gains of 18.1\%--60.0\%. Dedicated trackers remain stronger on some localization or visibility measures, while LPA-CWM leads in completeness and continuity and achieves the best results on all main CMC measures on Kinetics. Within the CWM family, LPA-CWM also leads all TAP-Vid First measures across the three datasets.

In summary, our main contributions are:
\begin{itemize}
    \item We identify a key principle for CWM inference: physical priors should guide candidate reliability estimation and counterfactual response aggregation.
    \item We propose LPA-CWM, which uses a 3.0M-parameter adjudicator to predict candidate weights before windowed localization and one paired CWM re-evaluation.
    \item We introduce CMC for joint evaluation of localization, completeness, visibility, and continuity, demonstrating 18.1\%--60.0\% relative $\mathrm{DCA}_{\mathrm{avg}}$ gains over Uniform CWM.
\end{itemize}

\section{Related Work}

\paragraph{World models and counterfactual motion.}
Video world models learn predictive representations through feature prediction, as in V-JEPA~\citep{bardes2024vjepa}, or action-controllable generation, as in Genie~\citep{bruce2024genie}. CWM extracts correspondence by comparing factual and locally intervened predictions~\citep{bear2023cwm}, and later work applies this interface to physical dynamics~\citep{venkatesh2023physical}. Opt-CWM learns query-conditioned perturbations for flow and occlusion~\citep{kim2025motion}; KL-tracing measures the response of generative video models through KL divergence~\citep{kim2025kltracing}; and Point Prompting propagates colored markers through video diffusion models~\citep{shrivastava2026pointprompting}. LPA-CWM focuses on learning the relative reliability of the multiple responses already produced by a frozen CWM.

\paragraph{Diffusion models as structural priors.}
Diffusion features also support visual correspondence. SD4Match adapts Stable Diffusion features with image-pair-conditioned prompts~\citep{li2024sd4match}, while Track4Gen jointly trains video generation and point correspondence to reduce appearance drift~\citep{jeong2025track4gen}. NSG relates a diffusion score field to observed inter-frame changes as a spatiotemporal consistency signal~\citep{zhang2025nsg}. APA-CWM uses this signal for analytic candidate weighting, while LPA-CWM learns the weights from trajectory supervision.

\paragraph{Flow, correspondence, and point tracking.}
Our external baselines cover complementary motion interfaces. RAFT uses all-pairs correlations with iterative flow refinement~\citep{teed2020raft}, and SEA-RAFT improves the framework with probabilistic training and a more efficient design~\citep{wang2024sea}. SMURF learns optical flow through full-image warping and multi-frame self-supervision~\citep{stone2021smurf}. GMRW learns global correspondence with contrastive random walks~\citep{shrivastava2024gmrw}, whereas DODUO combines semantic priors with self-supervised flow~\citep{jiang2023doduo}. CoTracker3 is a dedicated point tracker trained with pseudo-labels from real videos~\citep{karaev2025cotracker3}; AllTracker extends dense long-range tracking with multi-frame flow and temporal attention~\citep{harley2025alltracker}.

\paragraph{Evaluation of motion consistency.}
General video and world-model benchmarks cover visual quality, temporal consistency, physical faithfulness, controllability, embodied utility, and counterfactual consistency~\citep{huang2024vbench,zheng2025vbench2,duan2025worldscore,shang2026worldarena,begiristain2026cronos}. Motion-specific evaluation includes FVMD for point-track motion distributions~\citep{liu2024fvmd}, Physics-IQ for prediction against real physical experiments~\citep{motamed2026physicsiq}, and TAPVid-3D for long-range 3D correspondence and visibility~\citep{koppula2024tapvid3d}. ITTO further diagnoses point trackers along motion-complexity and reappearance axes, exposing failures under fast motion and repeated occlusion~\citep{demler2025itto}. In contrast, CMC evaluates the quality of recovered 2D dynamic trajectories themselves, jointly measuring localization, completeness, visibility, and continuity while counting missing predictions on visible points as failures.

\section{Method}
\label{sec:method}

\begin{figure}[!t]
\centering
\includegraphics[width=\linewidth]{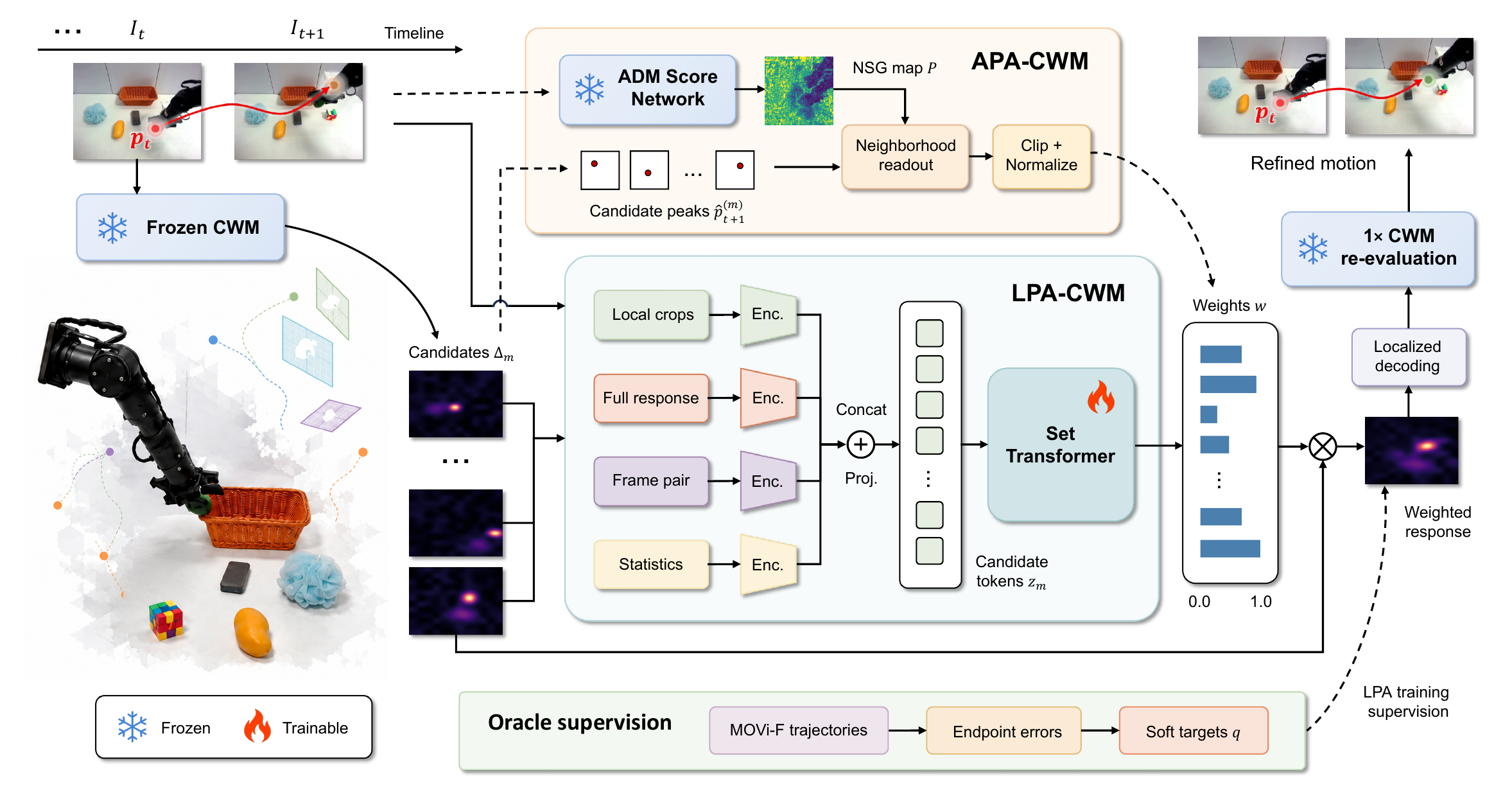}
\vspace{-0.6cm}
\caption{\textbf{Overview of APA-CWM and LPA-CWM.} Given a source--target RGB frame pair $(\mathbf{I}_t,\mathbf{I}_{t+1})$ and a query point $\mathbf{p}_t$, a frozen CWM generates motion candidates $\{\boldsymbol{\Delta}_m\}$. APA-CWM derives candidate weights from the NSG-derived map $\mathcal{P}$ around candidate peaks $\widehat{\mathbf{p}}_{t+1}^{(m)}$. LPA-CWM instead encodes local crops, full responses, the frame pair, and response statistics; the resulting tokens $\mathbf{z}_m$ are processed by a trainable Set Transformer to predict weights. The weighted response undergoes windowed localization and one CWM re-evaluation to obtain refined motion. During LPA training, endpoint errors from MOVi-F trajectories are converted into soft targets $\mathbf{q}$.}
\label{fig:method}
\end{figure}

Given a source--target frame pair and a query point, our method first uses a frozen
counterfactual world model (CWM) to produce multiple motion candidates. We then
compare two weighting schemes: an analytic rule, APA-CWM, and a
learned adjudicator, LPA-CWM. Both variants combine the candidate response maps
before applying the same localization and re-evaluation procedure.
Figure~\ref{fig:method} summarizes the method. 
% Only the 3.0M-parameter LPA is trained; the CWM and all decoding operations remain fixed.

\subsection{Counterfactual Motion Candidates from a Frozen World Model}

Let $\mathbf{I}_t,\mathbf{I}_{t+1}\in[0,1]^{256\times256\times3}$ denote a source--target RGB frame pair, and let $\mathbf{p}_t\in\Omega$ be a query in the source frame, where $\Omega$ is the $256\times256$ grid. We build on counterfactual world models~\citep{bear2023cwm,venkatesh2023physical,kim2025motion},
with an RGB predictor $\Psi^{\mathrm{RGB}}$ and a query-conditioned
intervention generator $\boldsymbol{\delta}_\theta$. Both
$\Psi^{\mathrm{RGB}}$ and $\boldsymbol{\delta}_\theta$ are pretrained and
remain fixed.

The generator produces the intervention
$\boldsymbol{\delta}_t=\boldsymbol{\delta}_\theta(\mathbf{I}_t,\mathbf{p}_t)
\in\mathbb{R}^{256\times256\times3}$. For each target-frame mask
$\mathcal{M}_m$, CWM measures how this intervention changes the target
prediction:
\begin{equation}
\boldsymbol{\Delta}_m=
\Psi^{\mathrm{RGB}}\!\left(
\mathbf{I}_t+\boldsymbol{\delta}_t,
\mathcal{M}_m(\mathbf{I}_{t+1})
\right)
-
\Psi^{\mathrm{RGB}}\!\left(
\mathbf{I}_t,
\mathcal{M}_m(\mathbf{I}_{t+1})
\right).
\end{equation}
Thus, $\boldsymbol{\Delta}_m(\mathbf{u})$ measures how much the prediction at
target pixel $\mathbf{u}$ changes when the queried source region is perturbed. A large
response indicates a strong dependency between that target location and the
queried source region. Repeating this process for $M$ masks produces
\begin{equation}
\mathcal{C}_t=
\left\{
\boldsymbol{\Delta}_m,
\widehat{\mathbf{p}}_{t+1}^{(m)}
\right\}_{m=1}^{M},
\qquad
\widehat{\mathbf{p}}_{t+1}^{(m)}
=
\arg\max_{\mathbf{u}\in\Omega}
\left\|
\boldsymbol{\Delta}_m(\mathbf{u})
\right\|_1 .
\end{equation}
Here, $\mathbf{u}$ indexes target-frame pixels and the $\ell_1$ norm reduces
the three RGB channels. The peak
$\widehat{\mathbf{p}}_{t+1}^{(m)}$ is the target location with the strongest
response under mask $m$, and therefore serves as that mask's proposed endpoint
for the query $\mathbf{p}_t$. Because different masks expose different target
contexts, their response maps and candidate endpoints can disagree. Uniform
CWM averages all responses equally, whereas APA-CWM and LPA-CWM estimate
candidate-specific weights.

\subsection{Analytic Candidate Weighting}

APA-CWM provides a closed-form reference based on the Normalized
Spatiotemporal Gradient (NSG)~\citep{zhang2025nsg}. Let
$\mathbf{x}_t,\mathbf{x}_{t+1}\in[-1,1]^{224\times224\times3}$ denote the
frame pair $(\mathbf{I}_t,\mathbf{I}_{t+1})$ after preprocessing for a frozen
Ablated Diffusion Model (ADM)~\citep{dhariwal2021adm}. The notation
distinguishes these ADM inputs from the $256\times256$ RGB frames used by CWM.
Let $\mathbf{s}_\phi$ be the frozen ADM score network, $t_s$ its diffusion
time, and $\mathbf{u}$ a location on the ADM grid.

Following NSG, we compute a \textbf{spatial score} $\mathcal{P}$ that compares the local ADM score magnitude with its alignment to the observed frame change:
\begin{equation}
\mathcal{P}(\mathbf{u})
=
\frac{
\left\|
\mathbf{s}_\phi(\mathbf{x}_t,t_s)(\mathbf{u})
\right\|_2
}{
\left|
\left\langle
\mathbf{s}_\phi(\mathbf{x}_t,t_s)(\mathbf{u}),
\mathbf{x}_{t+1}(\mathbf{u})-\mathbf{x}_t(\mathbf{u})
\right\rangle
\right|
+10^{-6}
}.
\end{equation}
The numerator is the local score magnitude, while the denominator is the absolute score--change inner product plus a $10^{-6}$ numerical stabilizer. The norm and inner product are computed over RGB channels. APA uses the resulting $\mathcal{P}$ map as an NSG-derived heuristic for candidate weighting, rather than a calibrated probability of correct correspondence.

We bilinearly resize $\mathcal{P}$ to the CWM response grid and keep the same
notation for the aligned map. For each candidate, APA averages this score in
the radius-six neighborhood $\mathcal{N}_6$ around its response peak and then
normalizes the candidate scores:
\begin{equation}
r_m^{\mathrm{APA}}
=
\frac{
1
}{
\left|
\mathcal{N}_6(\widehat{\mathbf{p}}_{t+1}^{(m)})
\right|
}
\sum_{\mathbf{u}\in
\mathcal{N}_6(\widehat{\mathbf{p}}_{t+1}^{(m)})}
\mathcal{P}(\mathbf{u}),
\qquad
w_m^{\mathrm{APA}}
=
\frac{
\operatorname{clip}(r_m^{\mathrm{APA}},10^{-3},10^3)
}{
\sum_j\operatorname{clip}(r_j^{\mathrm{APA}},10^{-3},10^3)
}.
\end{equation}
Here, $r_m^{\mathrm{APA}}$ is the local NSG score for candidate $m$. Clipping limits extreme values, and normalization converts these scores into nonnegative candidate weights $w_m^{\mathrm{APA}}$ that sum to one. APA thus provides a training-free way to favor candidates supported by the NSG cue. Since this cue is not directly supervised by endpoint accuracy, we next learn candidate reliability from dense motion trajectories.
% LPA-CWM does not use the diffusion score map.

\subsection{Learned Physical Adjudicator}
\label{sec:lpa}

\paragraph{Candidate representation and set reasoning.}
LPA represents each candidate using four sources of information.
\textbf{(1) Local visual evidence:} the local branch receives a 10-channel
$32\times32$ tensor containing source and target RGB crops, their absolute RGB
difference, and a crop of the candidate response. These inputs correspond to
$64\times64$ regions centered at $\mathbf{p}_t$ and
$\widehat{\mathbf{p}}_{t+1}^{(m)}$.
Before cropping, the RGB response is reduced and normalized as
$\widetilde d_m(\mathbf{u})
=
\|\boldsymbol{\Delta}_m(\mathbf{u})\|_1/
(\max_{\mathbf{v}}\|\boldsymbol{\Delta}_m(\mathbf{v})\|_1+10^{-8})$.
\textbf{(2) Full response:} a second branch encodes the complete
$\widetilde d_m$ map to retain the global shape and ambiguity of the response.
\textbf{(3) Frame context:} a third branch encodes the full frame pair to
provide scene-level visual information shared by all candidates.
\textbf{(4) Candidate statistics:} the final branch encodes 16 scalar
descriptors covering displacement, response concentration, ambiguity,
photometric change, and edge strength. The exact descriptors and encoder
architectures are given in Appendix~\ref{app:Implementation Details}.

The four encoded features are combined and projected to a 256-dimensional candidate token $\mathbf{z}_m$. We denote the Transformer encoder~\citep{lee2019settransformer} without candidate-index positional embeddings by $\operatorname{SetEnc}$. It compares all candidate tokens, after which a shared scoring head $h_{\mathrm{score}}$ produces one score per candidate:
\vspace{-0.1cm}
\begin{equation}
\left[
\widetilde{\mathbf{z}}_1,\ldots,\widetilde{\mathbf{z}}_M
\right]
=
\operatorname{SetEnc}
\left(
[\mathbf{z}_1,\ldots,\mathbf{z}_M]
\right),
\qquad
w_m^{\mathrm{LPA}}
=
\frac{
\exp(h_{\mathrm{score}}(\widetilde{\mathbf{z}}_m))
}{
\sum_j\exp(h_{\mathrm{score}}(\widetilde{\mathbf{z}}_j))
}.
\end{equation}
The softmax produces the candidate-weight vector
$\mathbf{w}^{\mathrm{LPA}}
=(w_1^{\mathrm{LPA}},\ldots,w_M^{\mathrm{LPA}})$,
whose nonnegative entries sum to one.
% The softmax converts these scores into normalized LPA weights $w_m^{\mathrm{LPA}}$. 
Without candidate-index positional embeddings, permuting the input candidates only permutes their output weights, while the weighted response remains unchanged. We also randomize candidate order during training.

\paragraph{Dense-trajectory supervision.}
We train LPA on candidate sets of size $M=10$ generated from adjacent MOVi-F frames~\citep{greff2022kubric}; training and validation videos are disjoint.
MOVi-F provides a dense ground-truth forward motion field
$\mathbf{F}_{t\rightarrow t+1}$, where
$\mathbf{F}_{t\rightarrow t+1}(\mathbf{p}_t)$ is the displacement of
$\mathbf{p}_t$ from frame $t$ to $t+1$. The corresponding \textbf{ground-truth endpoint} is
$\mathbf{p}_{t+1}^*
=
\mathbf{p}_t+\mathbf{F}_{t\rightarrow t+1}(\mathbf{p}_t)$.

We use this endpoint to measure how reliable each candidate is. Specifically,
$e_m$ is the endpoint error of candidate $m$, and these errors are converted
into a \textbf{soft reliability target} $\mathbf{q}=(q_1,\ldots,q_M)$:
\begin{equation}
e_m
=
\left\|
\widehat{\mathbf{p}}_{t+1}^{(m)}
-
\mathbf{p}_{t+1}^*
\right\|_2,
\qquad
q_m
=
\frac{
\exp(-e_m/\tau_q)
}{
\sum_j\exp(-e_j/\tau_q)
},
\qquad
\tau_q=4 .
\end{equation}
Candidates closer to the ground-truth endpoint receive larger $q_m$. Unlike a hard label that
selects only the single best candidate, $\mathbf{q}$ retains the relative
quality of all candidates. The temperature $\tau_q$ controls how strongly the
target distribution favors candidates with lower endpoint error. 

\paragraph{Training objective.}
We train LPA with the following objective:
% \vspace{-0.2cm}
\begin{equation}
\mathcal{L}
=
D_{\mathrm{KL}}
\left(
\mathbf{q}\|
\mathbf{w}^{\mathrm{LPA}}
\right)
+
\lambda_1\operatorname{SmoothL1}\!\left(
\sum_m
w_m^{\mathrm{LPA}}
\widehat{\mathbf{p}}_{t+1}^{(m)},
\mathbf{p}_{t+1}^*
\right)
+
\lambda_2\operatorname{CE}\!\left(
\mathbf{a},
\arg\min_m e_m
\right),
\end{equation}
where $\mathbf{a}
=(h_{\mathrm{score}}(\widetilde{\mathbf{z}}_1),\ldots,
h_{\mathrm{score}}(\widetilde{\mathbf{z}}_M))
\in\mathbb{R}^{M}$ is the vector of pre-softmax scores.
% where $\mathbf{a}=\left\{h_{\mathrm{score}}(\widetilde{\mathbf{z}}_m)\right\}_{m=1}^{M}$ denotes the pre-softmax scores. 
The KL term matches the predicted weights to the soft reliability targets $\mathbf{q}$. The SmoothL1 term encourages the weighted candidate endpoint toward the ground-truth endpoint, while the cross-entropy term encourages the lowest-error candidate to receive the highest score. Here we set $\lambda_1=2$ and $\lambda_2=0.2$.

The weighted candidate endpoint in the SmoothL1 term is used only as a differentiable training signal. During inference, LPA does not directly average candidate peak coordinates; instead, its predicted weights are applied to the full CWM response maps, which are then decoded as described next. MOVi-F correspondence filtering and further training details are given in Appendix~\ref{app:Implementation Details}.

\subsection{Weighted Aggregation and Localized Decoding}

\paragraph{Weighted response aggregation.}
Let $\mathbf{w}=(w_1,\ldots,w_M)$ collect the aggregation weights, with $w_m=1/M$ for Uniform CWM, $w_m=w_m^{\mathrm{APA}}$ for APA-CWM, and $w_m=w_m^{\mathrm{LPA}}$ for LPA-CWM.
% Let $w_m$ denote the uniform, analytic, or learned candidate weight used by Uniform CWM, APA-CWM, or LPA-CWM, respectively. 
We combine the full candidate responses and reduce the three RGB channels to a scalar localization map:
\begin{equation}
\overline{\boldsymbol{\Delta}}_w
=
\sum_{m=1}^{M}
w_m\boldsymbol{\Delta}_m,
\qquad
\overline d_w(\mathbf{u})
=
\frac{1}{3}
\left\|
\overline{\boldsymbol{\Delta}}_w(\mathbf{u})
\right\|_1,
\qquad
w_m\geq0,\quad
\sum_m w_m=1 .
\end{equation}
Here, $\overline{\boldsymbol{\Delta}}_w$ is the weighted CWM response and $\overline d_w(\mathbf{u})$ measures its response strength at pixel $\mathbf{u}$. Larger values indicate locations more strongly supported by the aggregated candidates.

\paragraph{Windowed localization and re-evaluation.}
To locate the endpoint, we turn $\overline d_w$ into a normalized spatial distribution
$\pi^{(0)}(\mathbf{u})\propto\exp(\beta_\pi \overline d_w(\mathbf{u}))$, with
$\beta_\pi=200$, normalized over all $\mathbf{u}\in\Omega$. Thus,
$\pi^{(0)}(\mathbf{u})$ gives the relative support for pixel $\mathbf{u}$ as the
endpoint, while $\pi^{(k)}$ denotes this distribution after $k$ windowing updates.
Standard localization directly selects the peak of $\pi^{(0)}$; our windowed
localization instead applies two updates to suppress distant secondary peaks.

\begin{wrapfigure}{r}{0.38\textwidth}
\vspace{-0.4cm}
\centering
\includegraphics[width=\linewidth]{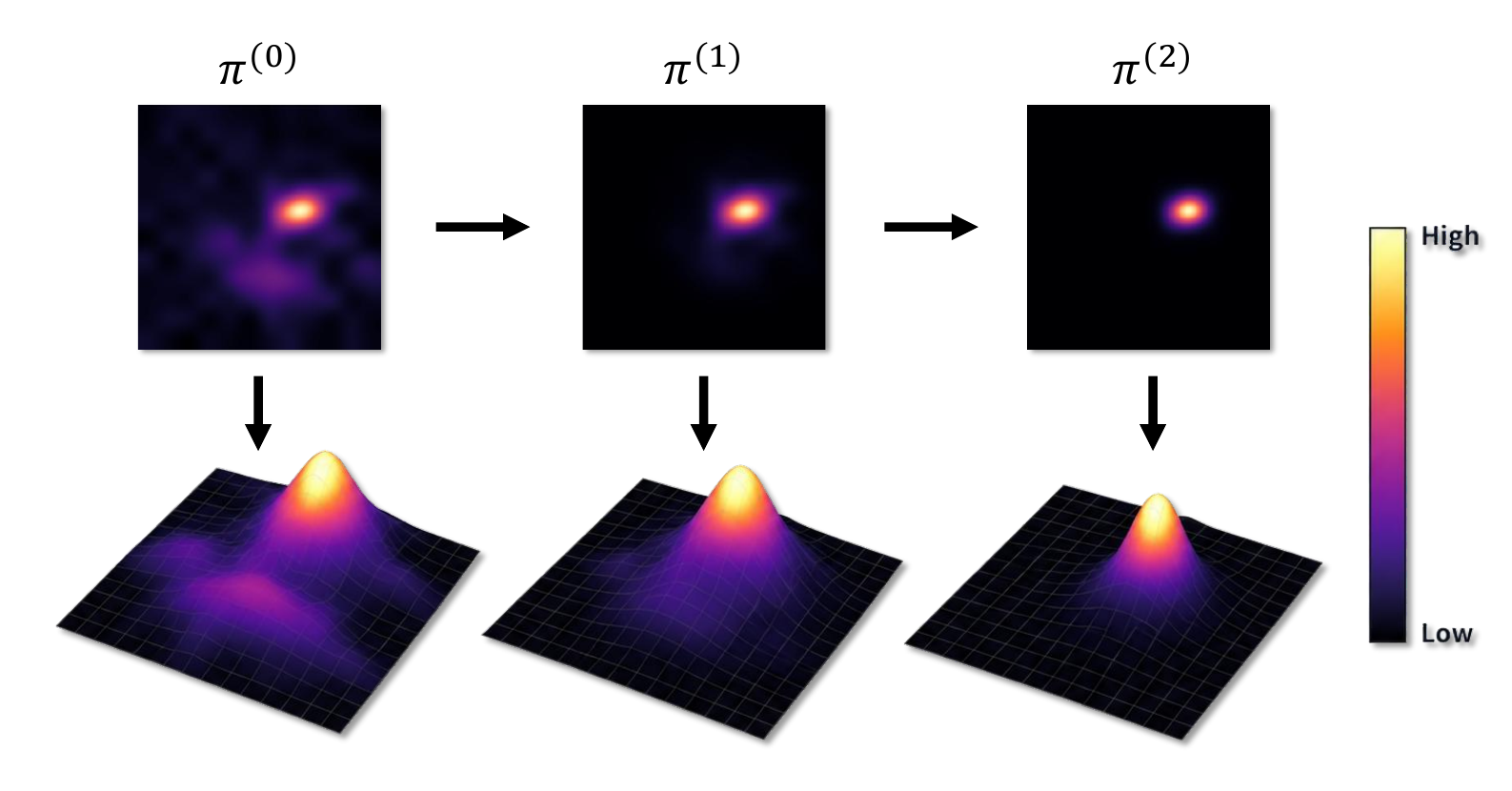}
\vspace{-0.7cm}
\caption{\textbf{Windowed localization progressively suppresses distant modes.}
Top: flattened maps from $\pi^{(0)}$ to $\pi^{(2)}$. Bottom: the corresponding 3D views.}
\label{fig:windowed_localization_demo}
\vspace{-1.5cm}
\end{wrapfigure}

At each update, we compute the spatial mean $\boldsymbol{\mu}^{(k)}$ and root-mean-square spatial spread $\sigma^{(k)}$ of the current distribution, then apply a Gaussian window centered at $\boldsymbol{\mu}^{(k)}$:
\begin{equation}
\begin{cases}
\boldsymbol{\mu}^{(k)}
=
\sum_{\mathbf{u}}
\pi^{(k)}(\mathbf{u})\mathbf{u},\\
\sigma^{(k)}
=
\left[
\sum_{\mathbf{u}}
\pi^{(k)}(\mathbf{u})
\left\|
\mathbf{u}-\boldsymbol{\mu}^{(k)}
\right\|_2^2
\right]^{1/2},
\\
\pi^{(k+1)}(\mathbf{u})
\propto
\pi^{(k)}(\mathbf{u})
\exp\!\left[
-
\frac{
\left\|
\mathbf{u}-\boldsymbol{\mu}^{(k)}
\right\|_2^2
}{
2[3\max(\sigma^{(k)},0.5)]^2
}
\right].
\end{cases}
\end{equation}
For $k=0,1$, we renormalize $\pi^{(k+1)}$ over $\Omega$ after applying the window. This reweighting downweights pixels farther from the current mean. The window width is $3\max(\sigma^{(k)},0.5)$ pixels, with the lower bound preventing window collapse. The discrete argmax of $\pi^{(2)}$ gives the initial endpoint.

Finally, this initial endpoint determines the target-mask center for one fresh, full-field CWM evaluation, while the source query remains $\mathbf{p}_t$. 
The new $M$ responses are averaged uniformly and decoded to obtain the refined endpoint $\widehat{\mathbf{p}}_{t+1}$; APA and LPA are not rerun, and their initial weights are not reused.
The pairwise motion is $\widehat{\boldsymbol{\phi}}_t=\widehat{\mathbf{p}}_{t+1}-\mathbf{p}_t$. In evaluation, each target is predicted from the fixed first-visible query without recursion. Visibility comes from the final paired CWM response-strength signal; LPA has no separate visibility head. Further details are provided in Appendix~\ref{app:Implementation Details}.

\section{Experiments}

Our experiments mainly address three questions: \textbf{(1)} Does learned candidate weighting improve motion recovered from a frozen CWM? \textbf{(2)} How does LPA-CWM compare with analytic weighting and external motion baselines under CMC and standard point-tracking evaluation? \textbf{(3)} Which components and weighting behaviors contribute to the improvement?

\subsection{Experimental Setup}

\paragraph{Models and data.}
Uniform CWM, APA-CWM, and LPA-CWM use the same frozen predictor, intervention generator, evaluation queries, and mask-sampling protocol. They differ in initial candidate weighting, while paired re-evaluation is performed separately at the endpoint produced by each variant. We fix $M=10$, use one paired re-evaluation ($R=1$), set $\beta_\pi=200$, and apply two windowed localization updates. Only the 3.0M-parameter LPA is trained on MOVi-F trajectories, with checkpoint selection performed on held-out MOVi-F videos. We evaluate on three datasets---DAVIS~\citep{ponttuset2017davis}, Kinetics~\citep{kay2017kinetics}, and RoboTAP~\citep{vecerik2024robotap}---covering 90 videos and approximately 1,300 annotated point trajectories in total. All methods use the same evaluation subsets and queries; further protocol details are provided in \hyperref[app:datasets-eval]{Appendix~\ref*{app:datasets-eval}}.

\textbf{Evaluation and comparisons.}
We evaluate motion recovery using our CMC protocol and the standard TAP-Vid First protocol. For TAP-Vid First, we report Average Jaccard (AJ), the average fraction of points within the benchmark distance thresholds ($<\delta^x_{\rm avg}$), mean and median endpoint distance (AD and MD), and occlusion accuracy (OA). AJ jointly reflects localization and visibility, while AD and MD summarize localization error. External baselines include SMURF, GMRW, DODUO, CoTracker3, RAFT, and SEA-RAFT, covering optical flow, correspondence, and dedicated point tracking. All methods receive identical evaluation frames and queries, and no baseline uses post-hoc smoothing or repair. Appendix~\ref{app:external-adaptation} describes the flow and correspondence adapters.

\begin{figure*}[!t]
\centering
\includegraphics[width=0.95\linewidth]{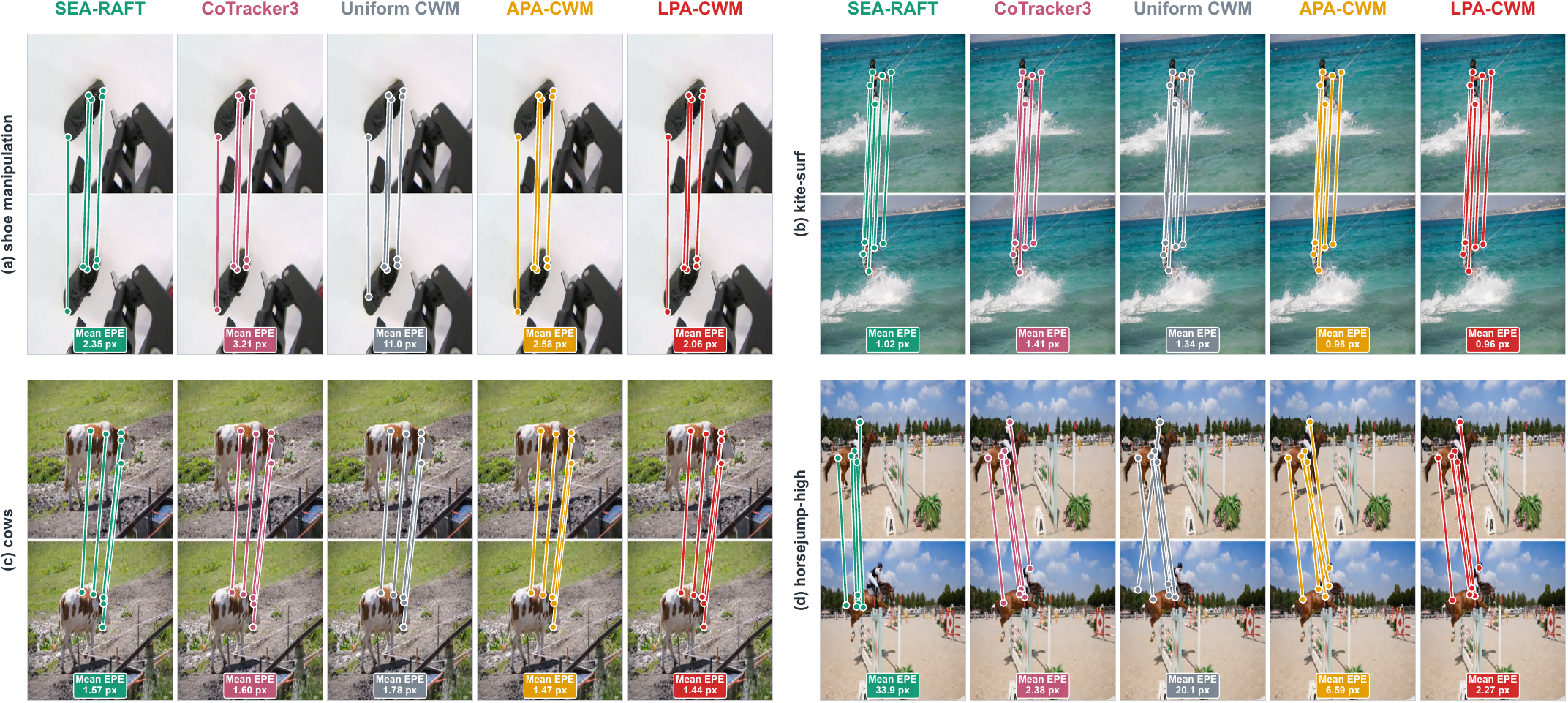}
\vspace{-0.25cm}
\caption{\textbf{Qualitative motion correspondence.} Five methods are compared on identical frames and visible queries across manipulation, human-action, and animal-motion scenes. Badges report mean EPE in pixels. LPA-CWM yields accurate, coherent correspondences across the displayed cases.}
\label{fig:qualitative}
\end{figure*}

\subsection{Completeness-aware Motion Correspondence}
\label{sec:cmc}

\begin{wrapfigure}{r}{0.25\linewidth}
\vspace{-0.70cm}
\centering
\includegraphics[width=0.84\linewidth]{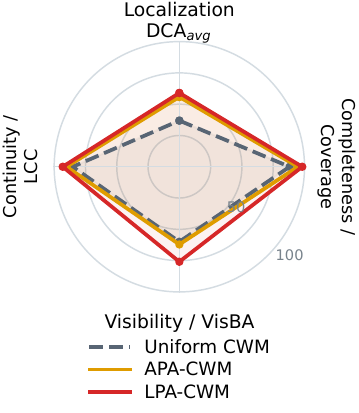}
\vspace{-0.15cm}
\caption{\textbf{CMC summary on DAVIS.} Higher is better.}
\label{fig:cmc-radar}
\vspace{-0.20cm}
\end{wrapfigure}

CMC evaluates post-query motion on ground-truth dynamic tracks, counting missing predictions on visible points as failures. We retain tracks with at least 4 pixels of cumulative post-query displacement at $256\times256$ resolution. CMC uses four dimensions. \textbf{1. Localization} uses DCA$_{\rm avg}$ (DCA), averaging visible-and-correct predictions over $\{1,2,4,8,16\}$-pixel thresholds. \textbf{2. Completeness} uses Coverage (Cov.) and Complete@0.8 (C@.8), \textbf{3. visibility} by visibility balanced accuracy (VisBA) and occlusion F1 (OF1), and \textbf{4. continuity} by LCC, the longest uninterrupted predicted-visible run. Figure~\ref{fig:cmc-radar} shows one metric per dimension; Table~\ref{tab:cross-dataset-main} reports the full set. Complete Track Success (CTS) measures track-level success by requiring at least 80\% coverage and mean endpoint error (EPE) at most 4 pixels on jointly visible points. Definitions, threshold sweeps, and fragmentation diagnostics appear in Appendix~\ref{app:cmc-definition}.

\begin{table*}[!t]
\centering
\vspace{-0.4cm}
\caption{\textbf{Cross-dataset evaluation on DAVIS, Kinetics, and RoboTAP.}
LPA is trained exclusively on MOVi-F and directly evaluated on all three datasets without adaptation. CMC metrics are percentages. For TAP-Vid First, AJ, $<\delta_{\mathrm{avg}}^{x}$, and OA are percentages, while AD and MD are pixels. Bold marks the best result among the three CWM variants. Stronger external results are shaded separately.}
\label{tab:cross-dataset-main}
\tablecaptionspace
\footnotesize
\setlength{\tabcolsep}{1.8pt}
\renewcommand{\arraystretch}{1.06}

\resizebox{0.97\linewidth}{!}{%
\begin{tabular}{c|l|ccccccc|ccccc}
\toprule
\toprule
\multicolumn{1}{c|}{\multirow{2}{*}{\textbf{Dataset}}}
& \multicolumn{1}{c|}{\multirow{2}{*}{\textbf{Method}}}
& \multicolumn{7}{c|}{\textbf{CMC (Ours)}}
& \multicolumn{5}{c}{\textbf{TAP-Vid First}} \\
\cmidrule(lr){3-9}\cmidrule(lr){10-14}
& & Cov.$\uparrow$ & C@.8$\uparrow$ & DCA$\uparrow$ & CTS$\uparrow$ & VisBA$\uparrow$ & OF1$\uparrow$ & LCC$\uparrow$ & AJ$\uparrow$ & $<\delta_{\mathrm{avg}}^{x}\uparrow$ & AD$\downarrow$ & MD$\downarrow$ & OA$\uparrow$ \\
\midrule

\multirow{9}{*}{\textbf{DAVIS}}
& SMURF & 91.98 & 85.54 & 11.25 & 2.67 & 56.64 & 14.92 & 83.29 & 7.39 & 14.34 & 32.34 & 30.06 & 78.14 \\
& GMRW & 37.02 & 20.62 & 23.92 & 17.00 & 63.24 & 45.22 & 35.46 & 21.85 & 41.55 & 20.59 & 13.38 & 50.77 \\
& DODUO & 31.44 & 15.04 & 22.25 & 14.42 & 58.94 & 45.83 & 28.44 & 26.15 & 52.49 & 8.79 & 2.90 & 49.46 \\
& CoTracker3 & 92.31 & 85.33 & \cellcolor{gray!15}71.81 & \cellcolor{gray!15}70.50 & \cellcolor{gray!15}76.43 & \cellcolor{gray!15}58.97 & 81.01 & \cellcolor{gray!15}62.90 & \cellcolor{gray!15}76.81 & \cellcolor{gray!15}4.20 & \cellcolor{gray!15}2.01 & \cellcolor{gray!15}86.11 \\
& RAFT & 68.74 & 49.58 & 49.00 & 31.77 & 70.16 & 41.66 & 62.02 & 37.08 & 58.79 & 15.83 & 5.60 & 69.86 \\
& SEA-RAFT & 57.76 & 31.25 & 43.01 & 27.43 & 65.78 & 39.56 & 54.07 & 40.72 & 51.84 & 19.17 & 7.78 & 65.22 \\
\rowcolor{blue!10}
\cellcolor{white} & Uniform CWM & 88.10 & 80.96 & 36.87 & 21.46 & 59.91 & 33.91 & 85.20 & 24.31 & 40.21 & 13.10 & 3.98 & 76.56 \\
\rowcolor{blue!10}
\cellcolor{white} & APA-CWM (Ours) & 94.74 & 89.98 & 55.79 & 47.91 & 62.07 & 38.48 & 90.13 & 41.73 & 57.35 & 10.27 & 2.94 & 83.47 \\
\rowcolor{blue!10}
\cellcolor{white} & \textbf{LPA-CWM (Ours)} & \textbf{98.21} & \textbf{95.63} & \textbf{58.98} & \textbf{57.54} & \textbf{75.89} & \textbf{48.14} & \textbf{93.18} & \textbf{45.29} & \textbf{63.92} & \textbf{6.73} & \textbf{2.07} & \textbf{84.18} \\

\midrule

\multirow{9}{*}{\textbf{Kinetics}}
& SMURF & 89.67 & 81.89 & 19.16 & 14.17 & 62.69 & 35.11 & 82.40 & 17.28 & 29.43 & 27.67 & 20.56 & 70.48 \\
& GMRW & 55.28 & 36.25 & 32.38 & 30.42 & 65.99 & 46.34 & 51.43 & 29.83 & 48.48 & 17.87 & 12.13 & 53.52 \\
& DODUO & 67.04 & 49.87 & 40.21 & 39.87 & 78.37 & 58.39 & 59.83 & 36.89 & 46.32 & 13.80 & 4.15 & 67.69 \\
& CoTracker3 & 87.06 & 79.67 & 50.15 & 46.61 & 78.00 & 58.56 & 80.41 & \cellcolor{gray!15}50.86 & \cellcolor{gray!15}63.21 & \cellcolor{gray!15}9.85 & 3.60 & 84.80 \\
& RAFT & 80.83 & 69.24 & 44.89 & 43.94 & 74.93 & 60.72 & 71.71 & 37.65 & 47.67 & 18.20 & 4.00 & 80.52 \\
& SEA-RAFT & 78.59 & 65.73 & 43.36 & 40.71 & 72.67 & 51.96 & 70.65 & 37.88 & 47.65 & 18.61 & 3.71 & 79.45 \\
\rowcolor{blue!10}
\cellcolor{white} & Uniform CWM & 87.14 & 77.11 & 40.26 & 34.54 & 67.66 & 44.97 & 79.95 & 35.44 & 46.62 & 15.26 & 7.44 & 85.99 \\
\rowcolor{blue!10}
\cellcolor{white} & APA-CWM (Ours) & 89.08 & 83.78 & 46.22 & 37.10 & 77.66 & 59.22 & 84.60 & 35.66 & 47.40 & 12.15 & 4.01 & 87.19 \\
\rowcolor{blue!10}
\cellcolor{white} & \textbf{LPA-CWM (Ours)} & \textbf{94.65} & \textbf{89.97} & \textbf{51.92} & \textbf{58.25} & \textbf{82.97} & \textbf{62.67} & \textbf{86.84} & \textbf{41.91} & \textbf{53.12} & \textbf{10.87} & \textbf{3.45} & \textbf{88.96} \\

\midrule

\multirow{9}{*}{\textbf{RoboTAP}}
& SMURF & 90.33 & 81.38 & 21.13 & 11.67 & 57.17 & 28.66 & 78.28 & 27.03 & 42.56 & 23.89 & 15.91 & 75.68 \\
& GMRW & 30.31 & 10.89 & 18.41 & 7.01 & 61.48 & 35.21 & 28.76 & 20.83 & 34.73 & 23.49 & 13.76 & 47.69 \\
& DODUO & 52.31 & 30.82 & 31.93 & 18.73 & 73.28 & 53.09 & 47.15 & 40.71 & 56.52 & 9.86 & 3.85 & 69.92 \\
& CoTracker3 & 81.02 & 73.92 & \cellcolor{gray!15}51.40 & 43.05 & 73.11 & 59.35 & 75.21 & \cellcolor{gray!15}56.58 & \cellcolor{gray!15}71.79 & 11.21 & \cellcolor{gray!15}1.96 & 80.14 \\
& RAFT & 76.69 & 54.71 & 49.10 & 41.27 & 70.60 & 44.27 & 69.46 & 45.57 & 57.69 & 13.06 & 3.21 & 72.00 \\
& SEA-RAFT & 75.58 & 62.14 & 42.45 & 24.68 & 65.23 & 37.86 & 64.72 & 47.38 & \cellcolor{gray!15}61.93 & 12.04 & \cellcolor{gray!15}2.61 & 77.62 \\
\rowcolor{blue!10}
\cellcolor{white} & Uniform CWM & 85.92 & 79.06 & 42.57 & 30.55 & 72.12 & 50.77 & 76.89 & 38.26 & 51.01 & 8.68 & 3.52 & 80.85 \\
\rowcolor{blue!10}
\cellcolor{white} & APA-CWM (Ours) & 91.10 & 81.78 & 46.91 & 42.34 & 75.63 & 57.44 & 78.71 & 43.17 & 56.90 & 7.88 & 2.88 & 83.24 \\
\rowcolor{blue!10}
\cellcolor{white} & \textbf{LPA-CWM (Ours)} & \textbf{92.00} & \textbf{86.88} & \textbf{50.26} & \textbf{54.22} & \textbf{78.03} & \textbf{61.00} & \textbf{85.49} & \textbf{47.42} & \textbf{58.35} & \textbf{7.74} & \textbf{2.75} & \textbf{86.45} \\

\bottomrule
\bottomrule
\end{tabular}}
\tablebottomspace
\end{table*}

Table~\ref{tab:cross-dataset-main} reports CMC results for all methods on three datasets. Within the CWM family, LPA-CWM leads all seven CMC measures across the three datasets. Dedicated trackers such as CoTracker3 remain stronger on some localization or visibility measures, while LPA-CWM leads in completeness and continuity. Notably, on Kinetics, LPA-CWM achieves the best result among all evaluated methods on every main CMC measure. Relative to Uniform CWM, learned adjudication improves DCA$_{\rm avg}$ by 22.11, 11.66, and 7.69 percentage points on DAVIS, Kinetics, and RoboTAP, respectively; relative to APA-CWM, the corresponding gains are 3.19, 5.70, and 3.35 points.

\subsection{TAP-Vid First and Qualitative Results}

TAP-Vid First queries each point at its first visible frame and evaluates the subsequent trajectory~\citep{doersch2022tapvid}. Within the CWM family, LPA-CWM achieves the best result on all main measures across the three datasets (Table~\ref{tab:cross-dataset-main}), while CoTracker3 remains stronger on several conventional tracking measures, particularly on DAVIS. Together with CMC, these results show that learned candidate weighting improves both tracking accuracy and motion completeness.

Figure~\ref{fig:qualitative} shows four qualitative comparisons spanning manipulation, human-action, and animal-motion scenes. LPA-CWM gives the lowest displayed mean EPE in all four cases and remains stable under large motion and appearance changes. Additional examples are shown in \hyperref[fig:additional-qualitative]{Appendix Figure~\ref*{fig:additional-qualitative}}.

\subsection{Ablation Studies}

\paragraph{Mask count and re-evaluation.}
We select $M$ and $R$ on 40 fixed MOVi-F query groups, sweeping $M\in\{5,10,20\}$ and $R\in\{0,1,2\}$. The best accuracy--cost setting is $M=10,R=1$; $M=20$ increases latency without improving accuracy, while $R=2$ does not further reduce EPE.

\paragraph{Architecture, objectives, and decoding.}
Table~\ref{tab:lpa-targeted-ablation} isolates the main design choices under matched inference settings. Removing candidate-set interaction or visual cues consistently reduces performance, while coordinate or best-candidate supervision improves over KL matching alone. The full objective performs best on all reported measures, and windowed localization improves every weighting rule. Appendix~\ref{app:ablations} provides the full sweeps, timing results, and detailed controls.

\begin{figure*}[!t]
\centering

\subfigure[Candidate-space weighted-coordinate EPE.]{
    \label{fig:lpa-mechanism-epe}
    \includegraphics[width=0.41\linewidth]{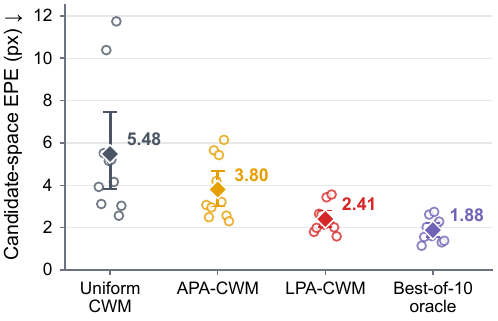}
}
\hspace{0.03\linewidth}
\subfigure[Weight versus ground-truth candidate rank.]{
    \label{fig:lpa-mechanism-rank}
    \includegraphics[width=0.41\linewidth]{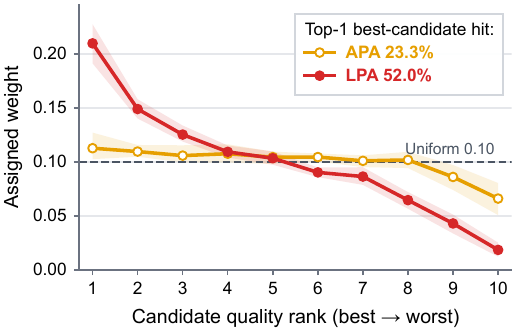}
}

\vspace{-0.2cm}
\caption{\textbf{Candidate weighting on MOVi-F validation videos.}
\textbf{(a)} Candidate-space weighted-coordinate EPE with video-macro means.
\textbf{(b)} Weight versus ground-truth candidate rank; the inset reports top-1 hit rates. Error bars and shading show 95\% video-cluster bootstrap intervals.}
\label{fig:lpa-mechanism-audit}
\end{figure*}

\paragraph{Candidate weighting.}
Figure~\ref{fig:lpa-mechanism-audit} shows that LPA reduces candidate-space EPE toward the best-of-10 oracle, favors better candidates, and more than doubles APA's top-1 hit rate. The rank-wise weight distribution shows that LPA assigns greater weight to more accurate candidates.

\begin{table*}[!tb]
\centering
\vspace{-0.4cm}
\caption{\textbf{Ablations on DAVIS under CMC and TAP-Vid First.} All configurations use $M=10$ and $R=1$. Architecture and objective variants use windowed localization; ``Std.'' and ``Win.'' denote standard and windowed localization. Metric conventions follow Table~\ref{tab:cross-dataset-main}; best values within each block are bold, enabling direct comparison across design choices.}
\label{tab:lpa-targeted-ablation}
\label{tab:components}
\tablecaptionspace
\scriptsize
\setlength{\tabcolsep}{1.8pt}
\renewcommand{\arraystretch}{0.96}
\resizebox{0.98\linewidth}{!}{%
\begin{tabular}{l|ccccccc|ccccc}
\toprule
\toprule
\multirow{2}{*}{\textbf{Configuration}}
& \multicolumn{7}{c|}{\textbf{CMC (Ours)}}
& \multicolumn{5}{c}{\textbf{TAP-Vid First}} \\
\cmidrule(lr){2-8}\cmidrule(lr){9-13}
& Cov.$\uparrow$ & C@.8$\uparrow$ & DCA$\uparrow$ & CTS$\uparrow$ & VisBA$\uparrow$ & OF1$\uparrow$ & LCC$\uparrow$
& AJ$\uparrow$ & $<\delta_{\mathrm{avg}}^{x}\uparrow$ & AD$\downarrow$ & MD$\downarrow$ & OA$\uparrow$ \\
\midrule
\multicolumn{13}{c}{\textbf{Architecture ablation}} \\
\midrule
Uniform CWM (reference) & 88.10 & 80.96 & 36.87 & 21.46 & 59.91 & 33.91 & 85.20 & 24.31 & 40.21 & 13.10 & 3.98 & 76.56 \\
w/o candidate-set interaction & 96.58 & 91.93 & 51.86 & 45.55 & 63.16 & 38.73 & 90.30 & 41.52 & 56.82 & 8.99 & 2.66 & 82.75 \\
Response + scalar cues only & 95.99 & 90.31 & 49.75 & 38.24 & 61.62 & 39.77 & 90.85 & 41.89 & 57.20 & 8.46 & 3.26 & 83.61 \\
Full LPA-CWM & \textbf{98.21} & \textbf{95.63} & \textbf{58.98} & \textbf{57.54} & \textbf{75.89} & \textbf{48.14} & \textbf{93.18} & \textbf{45.29} & \textbf{63.92} & \textbf{6.73} & \textbf{2.07} & \textbf{84.18} \\
\midrule
\multicolumn{13}{c}{\textbf{Objective ablation}} \\
\midrule
KL only & 95.73 & 91.78 & 49.64 & 36.99 & 60.47 & 35.03 & 90.77 & 41.56 & 57.48 & 9.09 & 2.80 & 82.84 \\
KL + coordinate supervision & 96.60 & 94.33 & 52.13 & 49.42 & 62.00 & 39.07 & 91.59 & 42.51 & 58.21 & 8.11 & 2.12 & 83.34 \\
KL + best-candidate supervision & 97.89 & 95.00 & 51.27 & 48.75 & 62.34 & 38.50 & 92.89 & 42.92 & 57.96 & 7.95 & 2.16 & 83.73 \\
Full objective & \textbf{98.21} & \textbf{95.63} & \textbf{58.98} & \textbf{57.54} & \textbf{75.89} & \textbf{48.14} & \textbf{93.18} & \textbf{45.29} & \textbf{63.92} & \textbf{6.73} & \textbf{2.07} & \textbf{84.18} \\
\midrule
\multicolumn{13}{c}{\textbf{Aggregation and localization}} \\
\midrule
Uniform + Std. & 86.93 & 77.00 & 34.54 & 17.71 & 57.14 & 31.43 & 84.05 & 23.25 & 39.46 & 13.61 & 4.26 & 76.18 \\
APA + Std. & 89.49 & 83.46 & 50.78 & 44.80 & 61.30 & 34.11 & 89.78 & 40.54 & 56.19 & 10.54 & 3.12 & 83.29 \\
LPA + Std. & 95.02 & 88.79 & 55.86 & 48.33 & 67.33 & 39.62 & 91.71 & 42.80 & 58.47 & 7.17 & 2.85 & 83.85 \\
Uniform + Win. & 88.10 & 80.96 & 36.87 & 21.46 & 59.91 & 33.91 & 85.20 & 24.31 & 40.21 & 13.10 & 3.98 & 76.56 \\
APA + Win. & 94.74 & 89.98 & 55.79 & 47.91 & 62.07 & 38.48 & 90.13 & 41.73 & 57.35 & 10.27 & 2.94 & 83.47 \\
LPA + Win. & \textbf{98.21} & \textbf{95.63} & \textbf{58.98} & \textbf{57.54} & \textbf{75.89} & \textbf{48.14} & \textbf{93.18} & \textbf{45.29} & \textbf{63.92} & \textbf{6.73} & \textbf{2.07} & \textbf{84.18} \\
\bottomrule
\bottomrule
\end{tabular}}
\tablebottomspace
\end{table*}

% \begin{figure*}[!t]
% \centering
% \includegraphics[width=0.85\linewidth]{figure/fig05.pdf}
% \vspace{-0.2cm}
% \caption{\textbf{Candidate weighting on MOVi-F validation videos.}
% \textbf{(a)} Candidate-space weighted-coordinate EPE with video-macro means.
% \textbf{(b)} Weight versus ground-truth candidate rank; the inset reports top-1 hit rates. Error bars and shading show 95\% video-cluster bootstrap intervals.}
% % \vspace{-0.2cm}
% \label{fig:lpa-mechanism-audit}
% \end{figure*}

\FloatBarrier

\section{Conclusion and Limitations}

We presented LPA-CWM for motion reasoning with a frozen counterfactual world model. LPA learns relative reliability across competing CWM candidates from dense trajectory supervision, while APA-CWM provides a training-free analytic reference. We also introduced CMC to evaluate localization, completeness, visibility, and continuity on dynamic tracks. Across DAVIS, Kinetics, and RoboTAP, LPA-CWM consistently improves over Uniform CWM and APA-CWM under both CMC and TAP-Vid First, with ablations confirming the contributions of candidate interaction, trajectory supervision, and windowed localization.

Three limitations remain. First, LPA can only reweight motion evidence already present in the frozen candidate set; missing or systematically incorrect motion requires improved candidate generation or adaptation of the underlying CWM. Second, LPA is trained on synthetic MOVi-F trajectories, while the current quantitative evaluation spans real-world and embodied datasets within one CWM family. Extending trajectory supervision to real-world data and evaluating across broader datasets and additional world-model architectures will better establish generalization. Third, multi-mask inference and paired re-evaluation increase computation. More efficient candidate selection, shared computation across masks, or fewer CWM evaluations could reduce this cost.

\clearpage
% ============================================================
% ICLR 2027 Statements
% File: sections/6_Checklist.tex
% ============================================================

\section*{AI Use Statement}

Generative AI tools were used as research assistants for manuscript drafting and language editing, literature and baseline discovery, research ideation and experimental planning, code preparation and debugging, and the organization and interpretation of experimental results. They also assisted with preparing supporting documentation. All AI-assisted code and experimental procedures were reviewed by the authors, who executed the experiments and verified the reported numerical results against the corresponding logs and evaluation artifacts. References, bibliographic information, and related-work claims were checked by the authors. The authors take full responsibility for the final manuscript, scientific claims, experimental results, code, and associated artifacts.

\section*{Reproducibility Statement}

Section~\ref{sec:method} describes counterfactual candidate construction, analytic and learned candidate weighting, the training objectives, and the localization and paired re-evaluation procedure. Appendix~\ref{app:Implementation Details} specifies the LPA architecture, candidate features, MOVi-F correspondence filtering, and optimization and inference settings. LPA is trained exclusively on MOVi-F with disjoint training and validation videos, while the CWM predictor and intervention generator remain frozen. Checkpoint selection uses held-out MOVi-F videos, and evaluation on DAVIS, Kinetics, and RoboTAP is performed without dataset-specific adaptation.

Appendices~\ref{app:datasets-eval}--\ref{app:external-adaptation} document the evaluation datasets, TAP-Vid First and CMC protocols, metric definitions, and the coordinate and visibility adapters used for external baselines. All methods use the same evaluation subsets and queries, while the CWM variants use the same
frozen candidate generator and mask-sampling protocol. Appendix~\ref{app:ablations} details the controlled ablations, mask-count and re-evaluation sweeps, candidate-weight analysis, and latency measurement protocol. An overview of the method and results is available on the project page.\footnote{\url{https://LPA-CWM.github.io}}

\section*{Ethics Statement}

This work studies the reliability of motion correspondence recovered from frozen counterfactual world models. We use synthetic MOVi-F trajectories for LPA training and existing DAVIS, Kinetics, and RoboTAP benchmarks for evaluation. Real-world videos may depict people, so downstream applications should respect privacy and avoid non-consensual monitoring. LPA reweights existing motion candidates and cannot guarantee correct trajectories when the underlying candidates are unreliable. Likewise, high completeness or continuity alone does not establish accurate correspondence or physical validity. Applications in safety-critical robotic systems therefore require independent validation and appropriate safeguards. The use and redistribution of datasets, pretrained models, and derived artifacts should comply with their respective licenses and usage conditions.

\FloatBarrier
\clearpage
\bibliographystyle{iclr2027_conference}
\bibliography{iclr2027_conference}
\clearpage

\appendix
\raggedbottom
\numberwithin{figure}{section}
\numberwithin{table}{section}
\renewcommand*{\theHfigure}{\thesection.\arabic{figure}}
\renewcommand*{\theHtable}{\thesection.\arabic{table}}

\section{Implementation Details}
\label{app:Implementation Details}

\paragraph{Frozen candidate generator.}
The counterfactual predictor and intervention generator remain fixed during
LPA training and inference. CWM operates internally at $256\times256$
resolution and uses a $32\times32$ patch-grid mask, where each mask element
corresponds to one $8\times8$-pixel image patch; approximately 89.9\% of the
target-frame patch grid is hidden. For clarity, Section~\ref{sec:method} uses the canonical $256\times256$ CWM grid for notation; in the implementation, responses and candidate coordinates are mapped to the evaluator grid $H\times W$ before LPA input construction. Its candidate endpoint is obtained from the peak of the channel-reduced absolute RGB response. Averaging rather than summing over RGB
channels changes only a constant factor and therefore does not change the
peak location.

\paragraph{LPA architecture.}
The complete LPA module contains 3,000,257 trainable parameters. The main text
uses height--width--channel notation, whereas the PyTorch implementation stores
image-like tensors in channel-first format. The local candidate encoder
therefore receives a $10\times32\times32$ tensor and applies four
$3\times3$ convolutional blocks with stride 2 and padding 1, using channel
dimensions $10\rightarrow32\rightarrow64\rightarrow128\rightarrow192$.
Each block uses GroupNorm and GELU, followed by global average pooling to
produce a 192-dimensional local feature.

Before global encoding, the complete normalized single-channel response is
bilinearly resized from the evaluator grid $H\times W$ to $32\times32$, and
the concatenated source--target RGB frames are bilinearly resized to
$64\times64$, both with \texttt{align\_corners=False}. The global response
encoder therefore receives a $1\times32\times32$ tensor and uses four
stride-2 convolutional blocks with channel dimensions
$1\rightarrow16\rightarrow32\rightarrow64\rightarrow96$, with $3\times3$
kernels, padding 1, GroupNorm, and GELU, followed by global average pooling to
produce a 96-dimensional feature. The shared frame-pair encoder receives the
resulting $6\times64\times64$ tensor and uses four analogous blocks with
channel dimensions $6\rightarrow32\rightarrow64\rightarrow128\rightarrow192$,
producing a 192-dimensional shared context feature.

The scalar encoder maps the 16-dimensional candidate descriptor through an MLP
with dimensions $16\rightarrow64\rightarrow64$, followed by GELU and
LayerNorm, to obtain a 64-dimensional feature. The four branches are
concatenated into a 544-dimensional representation and projected by
Linear$(544,256)$, LayerNorm, and GELU to form a 256-dimensional candidate
token. The set encoder contains four pre-normalized Transformer layers with
eight attention heads, a 512-dimensional feed-forward block, dropout 0.1, and
GELU activation. No candidate-index positional embedding is used. The shared
scoring head applies LayerNorm$(256)$, Linear$(256,128)$, GELU, and
Linear$(128,1)$, followed by a softmax over the candidate dimension.

\paragraph{Candidate statistics.}
The scalar branch receives the 16-dimensional descriptor listed in
Table~\ref{tab:lpa-scalar-statistics}. These statistics complement the learned
visual and response encoders with compact information about displacement,
response concentration, ambiguity, photometric change, and local edge
structure.

\begin{table}[H]
\centering
\caption{\textbf{Candidate scalar descriptor used by LPA.} The support column
indicates the spatial region used to compute each statistic.}
\label{tab:lpa-scalar-statistics}
\tablecaptionspace
\scriptsize
\setlength{\tabcolsep}{2.8pt}
\renewcommand{\arraystretch}{1.05}
\begin{tabularx}{\linewidth}{c l X l}
\toprule
Dim. & Statistic & Definition & Support \\
\midrule
1 & Normalized vertical displacement & Candidate $y$ displacement normalized by image height & Global \\
2 & Normalized horizontal displacement & Candidate $x$ displacement normalized by image width & Global \\
3 & Endpoint response & Response magnitude at the candidate endpoint & Single pixel \\
4 & Global response mean & Mean response magnitude & Full response \\
5 & Global response standard deviation & Population standard deviation of response magnitude & Full response \\
6 & Global response median & Median response magnitude & Full response \\
7 & Local peak ratio ($7\times7$) & Endpoint response divided by local mean response & $7\times7$ \\
8 & Local peak ratio ($15\times15$) & Endpoint response divided by local mean response & $15\times15$ \\
9 & Peak-to-median ratio & Endpoint response divided by global response median & Full response \\
10 & Peak-to-second-peak ratio & Endpoint response divided by the strongest response outside an $11\times11$ exclusion region & Global + local exclusion \\
11 & Local response mass ratio & Local response mass divided by total response mass & $7\times7$ \\
12 & Local response mass ratio & Local response mass divided by total response mass & $15\times15$ \\
13 & Local response mass ratio & Local response mass divided by total response mass & $31\times31$ \\
14 & Normalized spatial entropy & Entropy of the normalized spatial response distribution & Full response \\
15 & Local photometric change & Mean frame difference around the candidate endpoint & $7\times7$ \\
16 & Local source-edge strength & Mean source-frame edge magnitude around the candidate endpoint & $7\times7$ \\
\bottomrule
\end{tabularx}
\tablebottomspace
\end{table}

Dimensions 3--10 first undergo signed-log compression, using \(x\mapsto \operatorname{sign}(x)\log(1+|x|)\). All 16 dimensions are
then standardized using population means and standard deviations computed over
all candidate rows in the MOVi-F training split; the same normalization
statistics are stored with the checkpoint and reused at inference.
Candidate-centered scalar statistics use the rounded and image-clipped
candidate endpoint. Local windows are clipped to valid image support without
padding, and local means are normalized by the number of valid pixels.

\paragraph{Local patch sampling.}
Each local visual patch is centered at the source query or candidate endpoint,
covers a continuous $64\times64$ spatial extent on the evaluator grid, and is
bilinearly sampled onto a $32\times32$ grid with
\texttt{align\_corners=True}. If the sampling region extends beyond the image
domain, out-of-bounds values use border replication; candidate centers are
neither discarded nor recentered.

\paragraph{MOVi-F training correspondence filtering.}
Training samples are drawn from adjacent frames in the official MOVi-F
training split using the native $128\times128$ dense ground-truth forward and
backward flow fields. For a native-resolution source point $\mathbf{p}$, let
\[
\mathbf{p}'
=
\mathbf{p}
+
\mathbf{F}_{t\rightarrow t+1}(\mathbf{p}).
\]
The backward field is read at the rounded target coordinate, without bilinear
interpolation, and the forward--backward consistency error is
\[
e_{\mathrm{fb}}(\mathbf{p})
=
\left\|
\mathbf{F}_{t\rightarrow t+1}(\mathbf{p})
+
\mathbf{B}_{t+1\rightarrow t}
\left(\operatorname{round}(\mathbf{p}')\right)
\right\|_2.
\]
Here, $\mathbf{F}_{t\to t+1}$ and $\mathbf{B}_{t+1\to t}$
denote the ground-truth forward and backward flow fields
at the native $128\times128$ resolution, respectively.

A correspondence is retained only when $e_{\mathrm{fb}}\leq2.0$ pixels at the
native $128\times128$ resolution and both the source and warped target are at
least four pixels from the image boundary. Out-of-frame targets, points too
close to the boundary, and non-finite consistency errors are discarded rather
than clamped back into the valid domain. The sampling procedure additionally
requires a minimum motion magnitude of 0.25 pixels at the native resolution.
Accepted frames and coordinates are subsequently mapped to the common
$256\times256$ training scale.

\paragraph{LPA training.}
MOVi-F is partitioned at the source-video level before frame-pair and query
sampling, and each training sample retains the complete $M=10$ candidate set.
Candidate order is randomly permuted during training. AdamW uses a learning
rate of $3\times10^{-4}$, batch size 16, and a cosine schedule. Checkpoint
selection uses candidate-space weighted-coordinate EPE on held-out MOVi-F
validation videos. This criterion corresponds to the weighted-coordinate
supervision in Section~\ref{sec:lpa} and is distinct from the final deployed endpoint
EPE reported in the $M$--$R$ configuration study.

\paragraph{Analytic precursor.}
APA-CWM uses a frozen Ablated Diffusion Model (ADM) score network with
approximately 552M parameters, evaluated in FP16 at $224\times224$ resolution
and diffusion time $t_s=0.006$. The score map $P$ is computed once per frame
pair using the same numerical stabilizer $10^{-6}$ as in the main text and
bilinearly resampled to the aligned CWM response grid. APA averages the aligned statistic within a
radius-six neighborhood around each candidate response peak, clips the
candidate values to $[10^{-3},10^{3}]$, and applies L1 normalization across
candidates. The score map is used only by APA-CWM; LPA-CWM does not consume
the diffusion score map.

\paragraph{Counterfactual decoding.}
For each frame pair and query, $M$ denotes the number of mask-conditioned
candidate responses in the initial CWM set, whereas $R$ denotes the number of
paired re-evaluation passes after the initial localization. The two
hyperparameters therefore control different stages of inference.

When $R=0$, only the initial candidate set is weighted and localized, with no
additional CWM pass. For the default $R=1$, the decoder-specific initial
endpoint conditions one coarse-conditioned paired re-evaluation. The source
location remains the original query, while the initial endpoint determines
the center of the heuristic masking applied to the target frame. At
$256\times256$ resolution, both source and target retain complete spatial
support. A fresh set of $M$ masks is sampled, and the frozen CWM recomputes
$M$ mask-conditioned responses. APA and LPA are not rerun, the initial
candidate weights are not reused, and the fresh responses are uniformly
aggregated. The discrete peak of the resulting response distribution gives the
refined endpoint. Thus, the default $R=1$ stage is a full-field
coarse-conditioned paired re-evaluation rather than a local crop refinement.

\begin{figure}[!t]
\centering
\includegraphics[width=0.85\linewidth]{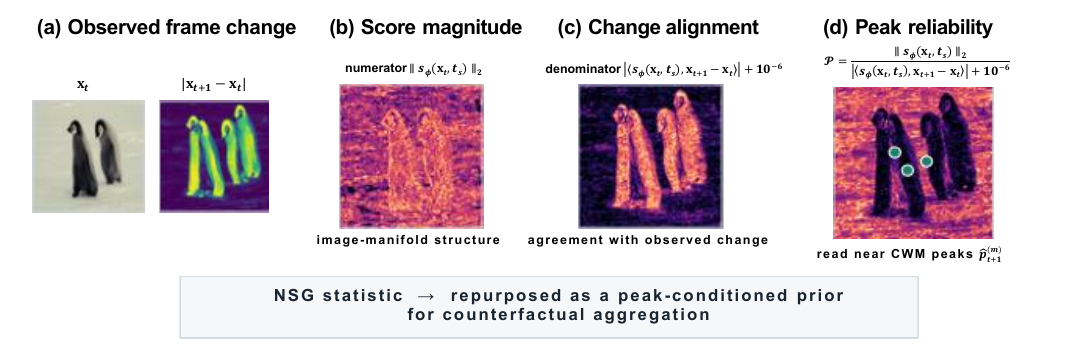}
\vspace{-0.3cm}
\caption{Evidence chain for the NSG-derived analytic precursor on a real frame
pair. The reference frame and observed change define the temporal evidence;
score magnitude and score--change alignment produce the analytic reliability
map $P$. APA-CWM averages this map within a local neighborhood around CWM candidate
peaks $\hat{p}_{t+1}^{(m)}$. LPA-CWM does
not use the diffusion score map; it learns candidate reliability from the
features described in Section~\ref{sec:lpa}. Heatmaps are independently normalized for
visualization.}
\label{fig:physical-prior}
\end{figure}

For $R>1$, later passes may become progressively local. The crop side follows $s_r=s_1\eta^{r-1}$ with fixed $\eta=0.75$; $\eta$ only controls crop size and is not learned or used in candidate weighting. For non-square inputs, the first square crop uses the shorter image dimension and is shifted to remain within the valid region rather than padded.
Here, $r=1,\ldots,R$ indexes the re-evaluation pass, $s_r$ is its crop side length, and $s_1$ is the first-pass crop side length, equal to 256 pixels for our square inputs.

\paragraph{Standard and windowed localization.}
Both decoders start from the same aggregated response distribution using the
softmax inverse-temperature parameter $\beta_{\pi}=200$. Standard
localization applies no Gaussian-window update and passes the discrete
argmax of the unwindowed response distribution to the subsequent $R$ stage.
Windowed localization performs two Gaussian-window reweighting steps. At each
step, the continuous distribution mean and spatial standard deviation determine
the window center and scale, with width
$3\max(\sigma,0.5)$; the distribution is renormalized after each update. The
coordinate passed to the $R$ stage is the discrete argmax after the second
update, not the continuous expectation. The two window updates operate within
a fixed response field and are distinct from the paired CWM re-evaluation
passes.

\paragraph{Trajectory construction.}
Under the reported TAP-Vid First and CMC evaluations, the first-visible point
remains the source query. For each evaluated target frame, CWM independently
processes the corresponding query--target pair, and the complete trajectory
is assembled from these direct predictions. Predictions at different target
times are not recursively propagated.

\paragraph{CWM visibility.}
LPA does not predict visibility. Let $\Delta_m^{\mathrm{final}}$ denote the three-channel response of mask $m$ from the final paired CWM pass. We compute
\[
s_m=\max_{\mathbf{u}}
\left\|\Delta_m^{\mathrm{final}}(\mathbf{u})\right\|_1,
\qquad
s=\frac{1}{M}\sum_{m=1}^{M}s_m,
\qquad
\hat{v}=\mathbb{1}[s\geq0.05].
\]
We retain the fixed response-strength threshold of $0.05$ used by the original CWM evaluator; it is not tuned on any evaluation dataset. This frozen CWM visibility rule is shared by Uniform CWM, APA-CWM, and LPA-CWM. The optional forward--backward occlusion refinement is disabled in all reported CWM experiments.

\FloatBarrier
\section{Datasets and Evaluation Protocols}
\label{app:datasets-eval}

\paragraph{Evaluation datasets.}
We evaluate TAP-Vid DAVIS and Kinetics~\citep{doersch2022tapvid} and
RoboTAP~\citep{vecerik2024robotap}, spanning real-world video and embodied
manipulation.
DAVIS emphasizes deformation, occlusion, distractors, and fast object motion,
whereas Kinetics contains diverse in-the-wild actions and camera motion.
RoboTAP extends the evaluation to robot-manipulation settings.
Across the three datasets, our evaluation covers 90 videos and nearly 1,300
annotated point trajectories in total. The same fixed evaluation subsets
and query tracks are used for all methods, and no model parameter is adapted
to any evaluation dataset.

\paragraph{Evaluation protocol.}
Following the TAP-Vid First protocol~\citep{doersch2022tapvid}, each query is
introduced at its first visible frame and evaluated on subsequent frames.
Ground-truth coordinates and visibility are taken directly from the
corresponding point-track annotations. The same query sets are used for
Uniform CWM, APA-CWM, LPA-CWM, and all external baselines, and missing
predictions on annotated visible points are retained as failures rather than
removed from evaluation. The CWM variants use the same locked videos, sampled
frames, first-visible queries, target frames, frozen candidate generator, and
mask-sampling protocol. Initial candidate weighting is method specific.
Because different weighting can produce different initial endpoints, the
paired re-evaluation is computed separately for each variant at its own
endpoint.

\paragraph{TAP-Vid metrics.}
We report Average Jaccard, points within the benchmark distance thresholds,
mean (AD) and median (MD) endpoint distance, and occlusion accuracy (OA). Average Jaccard (AJ)
jointly reflects localization and visibility, while the distance metrics
separately summarize typical and tail localization errors. CMC uses the same
ground-truth tracks but evaluates motion correspondence from the complementary
perspectives of localization, completeness, visibility, and temporal
continuity, as detailed in \hyperref[app:cmc-definition]{Appendix~\ref*{app:cmc-definition}}.

\FloatBarrier
\section{Completeness-aware Motion Correspondence}
\label{app:cmc-definition}

We introduce \emph{Completeness-aware Motion Correspondence} (CMC) to evaluate
motion correspondence on ground-truth dynamic tracks while explicitly
penalizing incomplete predictions. Dynamic tracks are selected from ground truth, and missing visible-point predictions remain in the accuracy denominator. The following metrics are computed separately for each video and then averaged with equal video weight.

For all three datasets, ground-truth coordinates and visibility labels are read
from the corresponding point-track annotations. We use TAP-Vid DAVIS and
Kinetics tracks for the two real-world video benchmarks and RoboTAP annotations for
RoboTAP. The same ground-truth trajectories are used by all compared methods.

Let $q_i$ be the query frame of trajectory $i$, $x^{\rm gt}_{i,t}$ and
$\hat x_{i,t}$ the ground-truth and predicted coordinates, and
$v^{\rm gt}_{i,t},\hat v_{i,t}\in\{0,1\}$ their visibility labels. For completeness and localization, we evaluate post-query, ground-truth-visible frames:
\[
E_i=\{t:t>q_i,\ v^{\rm gt}_{i,t}=1\}.
\]
The post-query ground-truth path length is
\[
L^{\rm gt}_{i,>q}=\sum_{t\ge q_i}
\mathbf{1}[v^{\rm gt}_{i,t}=v^{\rm gt}_{i,t+1}=1]
\lVert x^{\rm gt}_{i,t+1}-x^{\rm gt}_{i,t}\rVert_2 .
\]
The dynamic set $\mathcal D=\{i:L^{\rm gt}_{i,>q}\ge4\}$ is defined on a
common 256-pixel coordinate scale and is independent of all predictions.

\paragraph{Completeness.}
Per-track visibility coverage and its thresholded counterpart are
\[
C_i=\frac{\sum_{t\in E_i}\mathbf{1}[\hat v_{i,t}=1]}{|E_i|},\qquad
\mathrm{Coverage}=\frac{1}{|\mathcal D|}\sum_{i\in\mathcal D}C_i,
\]
\[
\mathrm{Complete@0.8}=\frac{1}{|\mathcal D|}\sum_{i\in\mathcal D}
\mathbf{1}[C_i\ge0.8].
\]
These are completeness diagnostics, not localization-accuracy measures.

\paragraph{Dynamic Correspondence Accuracy.}
With $e_{i,t}=\lVert\hat x_{i,t}-x^{\rm gt}_{i,t}\rVert_2$, we define
\[
\mathrm{DCA}_{\tau}=
\frac{\sum_{i\in\mathcal D}\sum_{t\in E_i}
\mathbf{1}[\hat v_{i,t}=1]\mathbf{1}[e_{i,t}<\tau]}
{\sum_{i\in\mathcal D}|E_i|}.
\]
Thus a prediction marked invisible contributes zero rather than disappearing
from the denominator. The primary scalar summary is
\[
\mathrm{DCA}_{\rm avg}=\frac{1}{5}
\sum_{\tau\in\{1,2,4,8,16\}}\mathrm{DCA}_{\tau}.
\]

\paragraph{Complete Track Success.}
Let $\bar e_i^{\rm joint}$ be the mean EPE over frames in $E_i$ that are also predicted visible. We set \(\bar e_i^{\mathrm{joint}}=+\infty\) when no jointly visible frame exists. The track-level criterion is
\[
\mathrm{CTS}(\tau,c)=\frac{1}{|\mathcal D|}\sum_{i\in\mathcal D}
\mathbf{1}[C_i\ge c]\mathbf{1}[\bar e_i^{\rm joint}\le\tau].
\]
We report CTS@(4,0.8) and CTS@(8,0.8). CTS is secondary to DCA because it still allows up to 20\% of a track to be absent.

\paragraph{Visibility and continuity.}
Visibility metrics use all post-query frames of the selected dynamic tracks, including ground-truth-occluded frames. Visibility balanced accuracy is
\[
\mathrm{VisBA}=\tfrac{1}{2}
(\mathrm{TPR}_{\rm visible}+\mathrm{TNR}_{\rm occluded}),
\]
and occlusion F1 is computed from post-query occlusion labels. 

For each track, longest continuous coverage (LCC) is the longest predicted-visible run within any ground-truth-visible segment, divided by $|E_i|$. 
% For each ground-truth-visible segment, longest continuous coverage (LCC) is the longest predicted-visible run divided by $|E_i|$. 
Fragmentation counts the number of $1\!\rightarrow\!0\!\rightarrow\!1$ visibility recoveries within such segments; Table~\ref{tab:cmc-visibility-full} reports the mean count.

\begin{table}[H]
\centering
\caption{\textbf{Full threshold-wise CMC accuracy results on DAVIS, Kinetics, and RoboTAP.}
All entries use the same 0--100 scale as Table~\ref{tab:cross-dataset-main}.
Cov. and C@.8 denote Coverage and Complete@0.8; CTS4 and CTS8 denote
CTS@(4,0.8) and CTS@(8,0.8), respectively. Results follow the presentation
convention of Table~\ref{tab:cross-dataset-main}.}
\label{tab:cmc-threshold-full}
\tablecaptionspace
\scriptsize
\setlength{\tabcolsep}{2.1pt}
\renewcommand{\arraystretch}{1.04}
\resizebox{0.98\linewidth}{!}{%
\begin{tabular}{c|l|cccccccccc}
\toprule
\multicolumn{1}{c|}{\textbf{Dataset}}
& \multicolumn{1}{c|}{\textbf{Method}}
& Cov.$\uparrow$
& C@.8$\uparrow$
& DCA@1$\uparrow$
& DCA@2$\uparrow$
& DCA@4$\uparrow$
& DCA@8$\uparrow$
& DCA@16$\uparrow$
& DCA$\uparrow$
& CTS4$\uparrow$
& CTS8$\uparrow$ \\
\midrule
\midrule

\multirow{9}{*}{\textbf{DAVIS}}
& SMURF & 91.98 & 85.54 & 1.72 & 4.11 & 8.72 & 14.07 & 27.62 & 11.25 & 2.67 & 3.33 \\
& GMRW & 37.02 & 20.62 & 11.47 & 21.20 & 26.10 & 28.57 & 32.25 & 23.92 & 17.00 & 17.00 \\
& DODUO & 31.44 & 15.04 & 6.76 & 18.23 & 25.89 & 29.59 & 30.79 & 22.25 & 14.42 & 15.04 \\
& CoTracker3 & 92.31 & 85.33
& \cellcolor{gray!15}40.55
& \cellcolor{gray!15}64.40
& \cellcolor{gray!15}80.49
& \cellcolor{gray!15}85.20
& \cellcolor{gray!15}88.41
& \cellcolor{gray!15}71.81
& \cellcolor{gray!15}70.50
& \cellcolor{gray!15}74.82 \\
& RAFT & 68.74 & 49.58 & \cellcolor{gray!15}29.93 & \cellcolor{gray!15}46.92 & 53.02 & 56.58 & 58.53 & 49.00 & 31.77 & 34.59 \\
& SEA-RAFT & 57.76 & 31.25 & \cellcolor{gray!15}27.93 & 40.57 & 47.35 & 49.01 & 50.18 & 43.01 & 27.43 & 30.54 \\
\rowcolor{blue!10}
\cellcolor{white} & Uniform CWM & 88.10 & 80.96 & 6.75 & 19.30 & 39.85 & 52.99 & 65.46 & 36.87 & 21.46 & 37.96 \\
\rowcolor{blue!10}
\cellcolor{white} & APA-CWM (Ours) & 94.74 & 89.98 & \textbf{19.31} & 41.71 & 61.91 & 73.94 & 82.05 & 55.79 & 47.91 & 58.14 \\
\rowcolor{blue!10}
\cellcolor{white} & \textbf{LPA-CWM (Ours)} & \textbf{98.21} & \textbf{95.63} & 17.87 & \textbf{44.73} & \textbf{67.49} & \textbf{80.13} & \textbf{84.67} & \textbf{58.98} & \textbf{57.54} & \textbf{66.79} \\

\midrule
\midrule

\multirow{9}{*}{\textbf{Kinetics}}
& SMURF & 89.67 & 81.89 & 4.61 & 9.36 & 16.87 & 24.29 & 40.65 & 19.16 & 14.17 & 14.79 \\
& GMRW & 55.28 & 36.25 & \cellcolor{gray!15}11.23 & 24.66 & 36.71 & 41.66 & 47.66 & 32.38 & 30.42 & 34.38 \\
& DODUO & 67.04 & 49.87 & 6.95 & 22.30 & 46.45 & 59.92 & 65.42 & 40.21 & 39.87 & 48.62 \\
& CoTracker3 & 87.06 & 79.67 & \cellcolor{gray!15}13.04 & 32.29 & 54.89 & 71.99 & 78.56 & 50.15 & 46.61 & 60.89 \\
& RAFT & 80.83 & 69.24 & \cellcolor{gray!15}10.02 & 26.02 & 49.49 & 67.38 & 71.55 & 44.89 & 43.94 & 60.75 \\
& SEA-RAFT & 78.59 & 65.73 & 9.49 & 25.19 & 47.29 & 65.01 & 69.80 & 43.36 & 40.71 & 56.55 \\
\rowcolor{blue!10}
\cellcolor{white} & Uniform CWM & 87.14 & 77.11 & 4.47 & 20.93 & 44.69 & 60.91 & 70.31 & 40.26 & 34.54 & 50.58 \\
\rowcolor{blue!10}
\cellcolor{white} & APA-CWM (Ours) & 89.08 & 83.78 & 8.44 & 22.86 & 47.03 & 72.16 & 80.61 & 46.22 & 37.10 & 64.66 \\
\rowcolor{blue!10}
\cellcolor{white} & \textbf{LPA-CWM (Ours)} & \textbf{94.65} & \textbf{89.97} & \textbf{9.58} & \textbf{35.22} & \textbf{61.37} & \textbf{72.47} & \textbf{80.97} & \textbf{51.92} & \textbf{58.25} & \textbf{67.04} \\

\midrule
\midrule

\multirow{9}{*}{\textbf{RoboTAP}}
& SMURF & 90.33 & 81.38 & \cellcolor{gray!15}7.53 & 16.12 & 22.03 & 26.82 & 33.15 & 21.13 & 11.67 & 16.25 \\
& GMRW & 30.31 & 10.89 & \cellcolor{gray!15}8.83 & 15.88 & 20.65 & 23.22 & 23.46 & 18.41 & 7.01 & 8.35 \\
& DODUO & 52.31 & 30.82 & \cellcolor{gray!15}8.51 & 23.62 & 34.39 & 44.86 & 48.28 & 31.93 & 18.73 & 29.57 \\
& CoTracker3 & 81.02 & 73.92
& \cellcolor{gray!15}26.55
& \cellcolor{gray!15}44.04
& 56.76 & 61.22 & 68.43
& \cellcolor{gray!15}51.40
& 43.05 & 49.70 \\
& RAFT & 76.69 & 54.71 & \cellcolor{gray!15}15.95 & \cellcolor{gray!15}40.72 & 58.73 & 62.72 & 67.37 & 49.10 & 41.27 & 46.64 \\
& SEA-RAFT & 75.58 & 62.14 & \cellcolor{gray!15}16.03 & \cellcolor{gray!15}36.57 & 49.31 & 52.86 & 57.48 & 42.45 & 24.68 & 25.31 \\
\rowcolor{blue!10}
\cellcolor{white} & Uniform CWM & 85.92 & 79.06 & \textbf{10.07} & 22.22 & 43.06 & 63.79 & 73.71 & 42.57 & 30.55 & 55.11 \\
\rowcolor{blue!10}
\cellcolor{white} & APA-CWM (Ours) & 91.10 & 81.78 & 6.33 & 21.73 & 51.54 & 75.91 & \textbf{79.05} & 46.91 & 42.34 & 59.53 \\
\rowcolor{blue!10}
\cellcolor{white} & \textbf{LPA-CWM (Ours)} & \textbf{92.00} & \textbf{86.88} & 6.47 & \textbf{29.41} & \textbf{59.94} & \textbf{76.56} & 78.93 & \textbf{50.26} & \textbf{54.22} & \textbf{66.00} \\
\bottomrule
\bottomrule
\end{tabular}}
\tablebottomspace
\end{table}

\begin{table}[H]
\centering
\vspace{-0.2cm}
\caption{\textbf{Visibility and continuity diagnostics for CMC on DAVIS, Kinetics, and RoboTAP.}
OF1 denotes occlusion F1. VisBA, OF1, and LCC are percentages; Frag. is the
mean fragmentation count $\times100$. Results follow the presentation
convention of Table~\ref{tab:cross-dataset-main}.}
\label{tab:cmc-visibility-full}
\tablecaptionspace
\scriptsize
\setlength{\tabcolsep}{3.0pt}
\renewcommand{\arraystretch}{1.08}
\resizebox{\linewidth}{!}{%
\begin{tabular}{l|cccc|cccc|cccc}
\toprule
\multicolumn{1}{c|}{\multirow{2}{*}{\textbf{Method}}}
& \multicolumn{4}{c|}{\textbf{DAVIS}}
& \multicolumn{4}{c|}{\textbf{Kinetics}}
& \multicolumn{4}{c}{\textbf{RoboTAP}} \\
\cmidrule(lr){2-5}
\cmidrule(lr){6-9}
\cmidrule(lr){10-13}
& VisBA$\uparrow$ & OF1$\uparrow$ & LCC$\uparrow$ & Frag.$\downarrow$
& VisBA$\uparrow$ & OF1$\uparrow$ & LCC$\uparrow$ & Frag.$\downarrow$
& VisBA$\uparrow$ & OF1$\uparrow$ & LCC$\uparrow$ & Frag.$\downarrow$ \\
\midrule
\midrule

SMURF
& 56.64 & 14.92 & 83.29 & 21.54
& 62.69 & 35.11 & 82.40 & 9.66
& 57.17 & 28.66 & 78.28 & 16.35 \\

GMRW
& 63.24 & 45.22 & 35.46 & 8.46
& 65.99 & 46.34 & 51.43 & 7.16
& 61.48 & 35.21 & 28.76 & 7.59 \\

DODUO
& 58.94 & 45.83 & 28.44 & 12.88
& 78.37 & 58.39 & 59.83 & 16.82
& 73.28 & 53.09 & 47.15 & 17.33 \\

CoTracker3
& \cellcolor{gray!15}76.43 & \cellcolor{gray!15}58.97 & 81.01 & 1.25
& 78.00 & 58.56 & 80.41 & 8.41
& 73.11 & 59.35 & 75.21 & 5.05 \\

RAFT
& 70.16 & 41.66 & 62.02 & 7.62
& 74.93 & 60.72 & 71.71 & 9.38
& 70.60 & 44.27 & 69.46 & 22.24 \\

SEA-RAFT
& 65.78 & 39.56 & 54.07 & 9.38
& 72.67 & 51.96 & 70.65 & 10.91
& 65.23 & 37.86 & 64.72 & 24.58 \\

\rowcolor{blue!10}
Uniform CWM
& 59.91 & 33.91 & 85.20 & 3.57
& 67.66 & 44.97 & 79.95 & 10.55
& 72.12 & 50.77 & 76.89 & 11.29 \\

\rowcolor{blue!10}
APA-CWM (Ours)
& 62.07 & 38.48 & 90.13 & 2.86
& 77.66 & 59.22 & 84.60 & 4.03
& 75.63 & 57.44 & 78.71 & 7.29 \\

\rowcolor{blue!10}
\textbf{LPA-CWM (Ours)}
& \textbf{75.89} & \textbf{48.14} & \textbf{93.18} & \textbf{0.63}
& \textbf{82.97} & \textbf{62.67} & \textbf{86.84} & \textbf{2.78}
& \textbf{78.03} & \textbf{61.00} & \textbf{85.49} & \textbf{2.59} \\

\bottomrule
\bottomrule
\end{tabular}}
\tablebottomspace
\vspace{0.1cm}
\end{table}

\begin{figure}[H]
\centering
\includegraphics[width=0.90\linewidth]{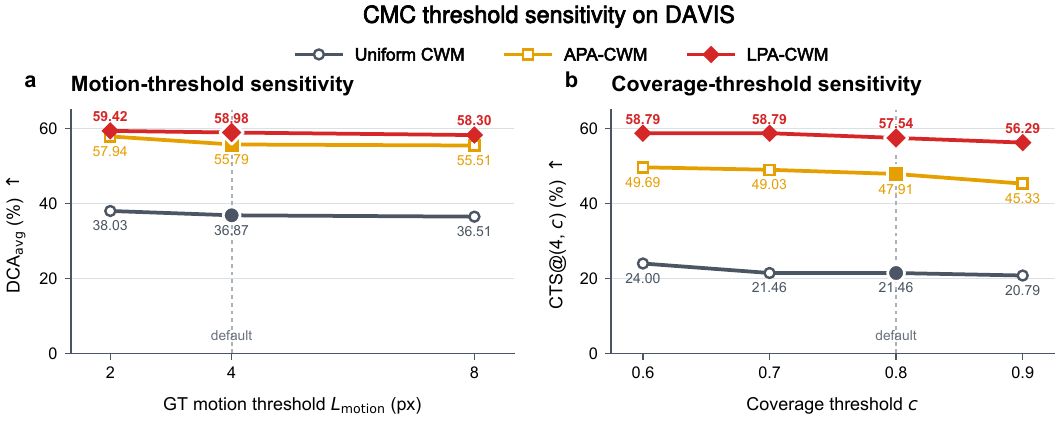}
\caption{\textbf{CMC threshold sensitivity on DAVIS.}
\textbf{(a)} DCA$_{\rm avg}$ as the dynamic-track motion threshold varies over $\{2,4,8\}$ pixels.
\textbf{(b)} CTS@(4,$c$) as the coverage threshold varies over $\{0.6,0.7,0.8,0.9\}$ with $L_{\rm motion}=4$ pixels.
Dashed vertical lines indicate the default settings.}
\label{fig:cmc-threshold-sensitivity}
\end{figure}

\paragraph{Threshold sensitivity.}
Figure~\ref{fig:cmc-threshold-sensitivity} evaluates whether the CMC conclusions depend on the default dynamic-motion and coverage thresholds. 
We vary the ground-truth motion threshold \(L_{\mathrm{motion}}\in\{2,4,8\}\) pixels and report DCA$_{\rm avg}$.
% With the coverage threshold fixed at $c=0.8$, we vary the ground-truth motion threshold $L_{\rm motion}\in\{2,4,8\}$ pixels and report DCA$_{\rm avg}$. 
LPA-CWM changes only from 59.42 to 58.30 across this range and remains above APA-CWM and Uniform CWM at every setting. We then fix $L_{\rm motion}=4$ pixels and vary the coverage threshold $c\in\{0.6,0.7,0.8,0.9\}$ for CTS@(4,$c$). LPA-CWM remains best throughout, decreasing moderately from 58.79 to 56.29 as the coverage requirement becomes stricter. The method ordering is therefore stable around the default $L_{\rm motion}=4$ pixels and $c=0.8$.

\FloatBarrier
\section{Adapting External Estimators}
\label{app:external-adaptation}

\paragraph{Temporal and coordinate adaptation.}
For the reported TAP-Vid First and CMC evaluations, the CWM variants and the
pairwise estimators RAFT, SEA-RAFT, DODUO, and SMURF use a direct
query-to-target construction. For a query introduced at frame $q$, each target
$t>q$ is predicted independently from the fixed first-visible query and the
corresponding pair $(I_q,I_t)$. The four pairwise baselines use their released
inference implementations and checkpoints within this adapter. CoTracker3
instead uses its official offline point-tracking interface, which jointly
processes the full video and the first-visible query, while GMRW uses its
released adjacent-frame tracking utility. Thus, all methods are evaluated on
the same query identities and target frames, while their native temporal
inference mechanisms remain method specific.

\paragraph{Visibility.}
Table~\ref{tab:external-visibility} summarizes the visibility rules used by
the external estimators. CoTracker3 uses the Boolean visibility returned by
its official offline predictor, which thresholds the native visibility score
at 0.9 and forces the annotated query point to be visible. No additional
forward--backward consistency test is applied.

\begin{table}[H]
\centering
\caption{\textbf{Visibility adaptation for external estimators.} Thresholds are
defined on the corresponding evaluator grid.}
\label{tab:external-visibility}
\tablecaptionspace
\scriptsize
\setlength{\tabcolsep}{2.8pt}
\renewcommand{\arraystretch}{1.04}
\begin{tabularx}{\linewidth}{l X c X}
\toprule
Method & Visibility source & Threshold & Sampling \\
\midrule
RAFT & Direct query-to-target forward--backward consistency & 6 px & Forward at integer query; backward bilinear \\
SEA-RAFT & Direct query-to-target forward--backward consistency & 6 px & Forward at integer query; backward bilinear \\
DODUO & Direct query-to-target forward--backward consistency & 6 px & Forward at integer query; backward bilinear \\
CoTracker3 & Native visibility from official offline predictor & 0.9 & Native sparse-query trajectory; no additional FB test \\
SMURF & Native occlusion map & 0.5 & Query-location lookup \\
GMRW & Official tracking utility with adjacent-frame forward--backward consistency & 3 px & Bilinear at propagated positions \\
\bottomrule
\end{tabularx}
\tablebottomspace
\end{table}

For the flow and correspondence estimators, out-of-frame or non-finite
predictions are treated as invisible by the corresponding adapters. CoTracker3
retains the native visibility returned by the official predictor without an
additional out-of-frame, non-finite, or forward--backward filter. These
evaluation-time visibility rules are distinct from the ground-truth
forward--backward filtering used to construct MOVi-F training samples in
Appendix~\ref{app:Implementation Details}.

\paragraph{Aggregation.}
All metrics are first computed from the same per-video query set and then
aggregated with equal video weight. No point or video is selected using a
method's predicted error, coverage, or visibility. Missing predictions enter
DCA as failures.

\FloatBarrier
\section{Ablation Protocols}
\label{app:ablations}

\paragraph{Joint $M$--$R$ sensitivity.}
The audit uses 40 query groups from 10 held-out MOVi-F videos, shared across configurations. We evaluate $(M,R)\in\{5,10,20\}\times\{0,1,2\}$. Masks are nested prefixes of the same ordered 20-mask set, avoiding independent draws for smaller $M$. All runs use full LPA-CWM on identical videos, frame pairs, and queries. We report final endpoint EPE after localization and $R$ paired re-evaluations. The sweep selects $M=10,R=1$ for the main comparison.

\begin{figure}[H]
\centering
\begin{minipage}[t]{0.49\linewidth}
    \centering
    \includegraphics[width=\linewidth]{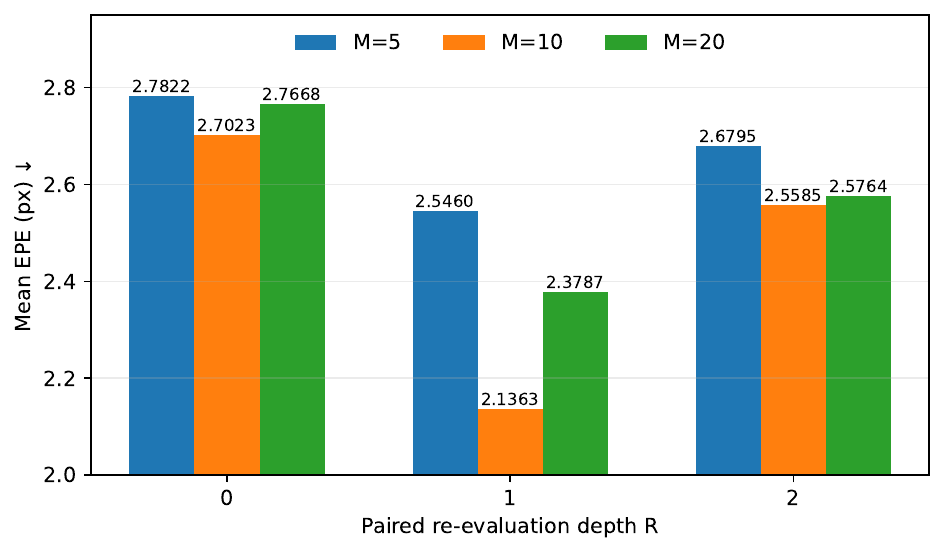}
    \vspace{-1mm}
    \textbf{(a) Final deployed endpoint EPE}
\end{minipage}
\hfill
\begin{minipage}[t]{0.49\linewidth}
    \centering
    \includegraphics[width=\linewidth]{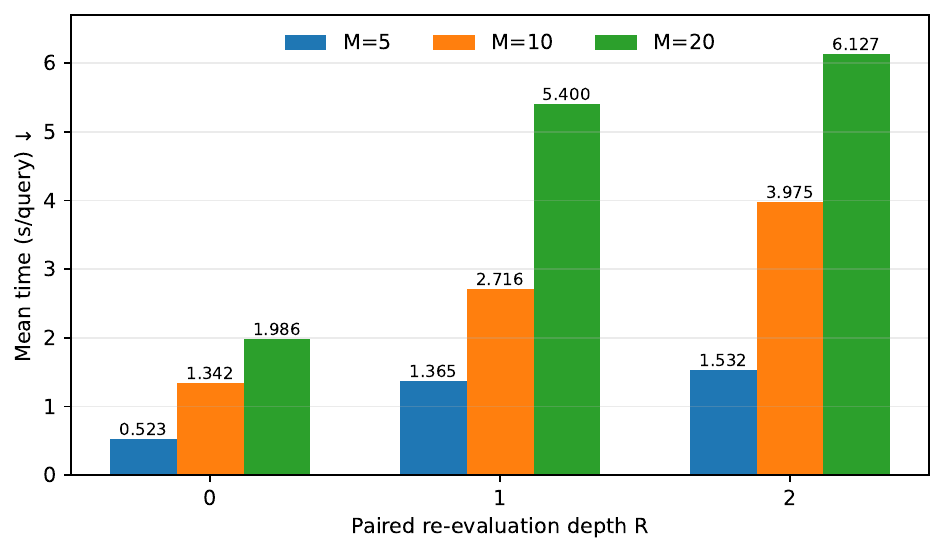}
    \vspace{-1mm}
    \textbf{(b) End-to-end inference time}
\end{minipage}
\caption{\textbf{Joint sensitivity to mask count and paired re-evaluation depth.}
Full LPA-CWM results on 40 fixed MOVi-F query groups.
\textbf{(a)} Final deployed endpoint EPE.
\textbf{(b)} End-to-end inference time per query.
$R=0$ omits paired re-evaluation; $M=10,R=1$ is the deployed setting.}
\label{fig:ablation-overview}
\label{fig:mask-refinement-sensitivity}
\end{figure}

\paragraph{Latency measurement.}
Figure~\ref{fig:ablation-overview} uses a single NVIDIA GeForce RTX 5090 with batch size 1, where each timed sample contains one adjacent frame pair and one query group. CUDA is synchronized immediately before and after the complete model call. Timing includes the factual frozen-CWM forward pass, $M$ mask-conditioned counterfactual evaluations, LPA feature extraction and set processing, candidate aggregation, localization, the specified $R$ paired CWM re-evaluation passes, and final endpoint extraction. Disk I/O, CPU preprocessing, host-to-device transfer, and model or checkpoint initialization are excluded. We therefore report end-to-end per-query latency under this protocol rather than throughput-optimized peak speed.

\paragraph{Candidate-weight analysis.}
Candidates are ranked within each complete ten-candidate group by ground-truth endpoint EPE. The top-1 oracle hit rate measures how often the highest-weight candidate is the minimum-EPE candidate. Effective support is $\exp(-\sum_m w_m\log w_m)$: it equals 1 for one-hot weights and 10 for uniform weights. Figure~\ref{fig:lpa-mechanism-audit} reports video-macro summaries. This analysis uses the MOVi-F validation split used for checkpoint selection and serves as mechanism evidence; real-video tracking is evaluated separately.

\paragraph{EPE definitions across analyses.}
Figure~\ref{fig:ablation-overview} evaluates 40 fixed query groups using the final deployed endpoint after localization and the requested $R$ re-evaluation passes; the default $M=10,R=1$ setting obtains 2.1363 pixels. Figure~\ref{fig:lpa-mechanism-audit} instead evaluates 640 complete $M=10$ candidate groups using the candidate-space weighted-coordinate surrogate before response-level localization and paired re-evaluation, where LPA obtains 2.41 pixels. These values are not directly comparable because they use different validation subsets and endpoint definitions.

\paragraph{Inference-stage ablation.}
All six configurations use the same frozen CWM, $M=10$, $R=1$, evaluation
seed, DAVIS batches, frame pairs, queries, and initial masks. The rows are
\emph{Uniform + Std.}, \emph{APA + Std.}, \emph{Uniform + Win.},
\emph{APA + Win.}, \emph{LPA + Std.}, and \emph{LPA + Win.} ``Std.'' uses
the discrete peak of the unwindowed aggregated response distribution as the
initial endpoint. ``Win.'' applies two Gaussian-window reweighting updates and
uses the discrete peak of the final distribution. Uniform, APA, and LPA
respectively denote equal, analytic, and learned weights for the initial
candidate set. All six rows then use the same one-pass coarse-conditioned
paired re-evaluation after their decoder-specific initial endpoint. The second
pass does not rerun APA or LPA, does not reuse the initial weights, and
uniformly aggregates the freshly generated responses. This $3\times2$ design
therefore separates initial candidate weighting from localization.

\paragraph{Learned architecture and objective ablations.}
Architecture variants use the complete three-term objective, the same windowed
localization, and the same $R=1$ paired re-evaluation. ``w/o candidate-set
interaction'' removes the permutation-equivariant set Transformer, while
``response + scalar cues only'' removes the local RGB and shared frame-pair raw
encoders but retains response maps, 16 scalar statistics, and candidate-set
interaction. 

Objective variants retain the full LPA architecture and compare
``KL only'', ``KL + coordinate supervision'',
``KL + best-candidate supervision'', and ``Full objective''. Table~\ref{tab:lpa-targeted-ablation} reports every row under both
CMC and TAP-Vid First on the same DAVIS evaluation set, queries, and initial mask protocol. The CMC columns summarize completeness, localization,
visibility, and continuity, while the TAP-Vid First columns report conventional
localization and occlusion measures. The inference procedure is fixed except for
the named architecture or objective component.

\FloatBarrier
\section{Additional Qualitative Results}

\begin{figure}[H]
\centering
\includegraphics[width=\textwidth,height=0.85\textheight,keepaspectratio]{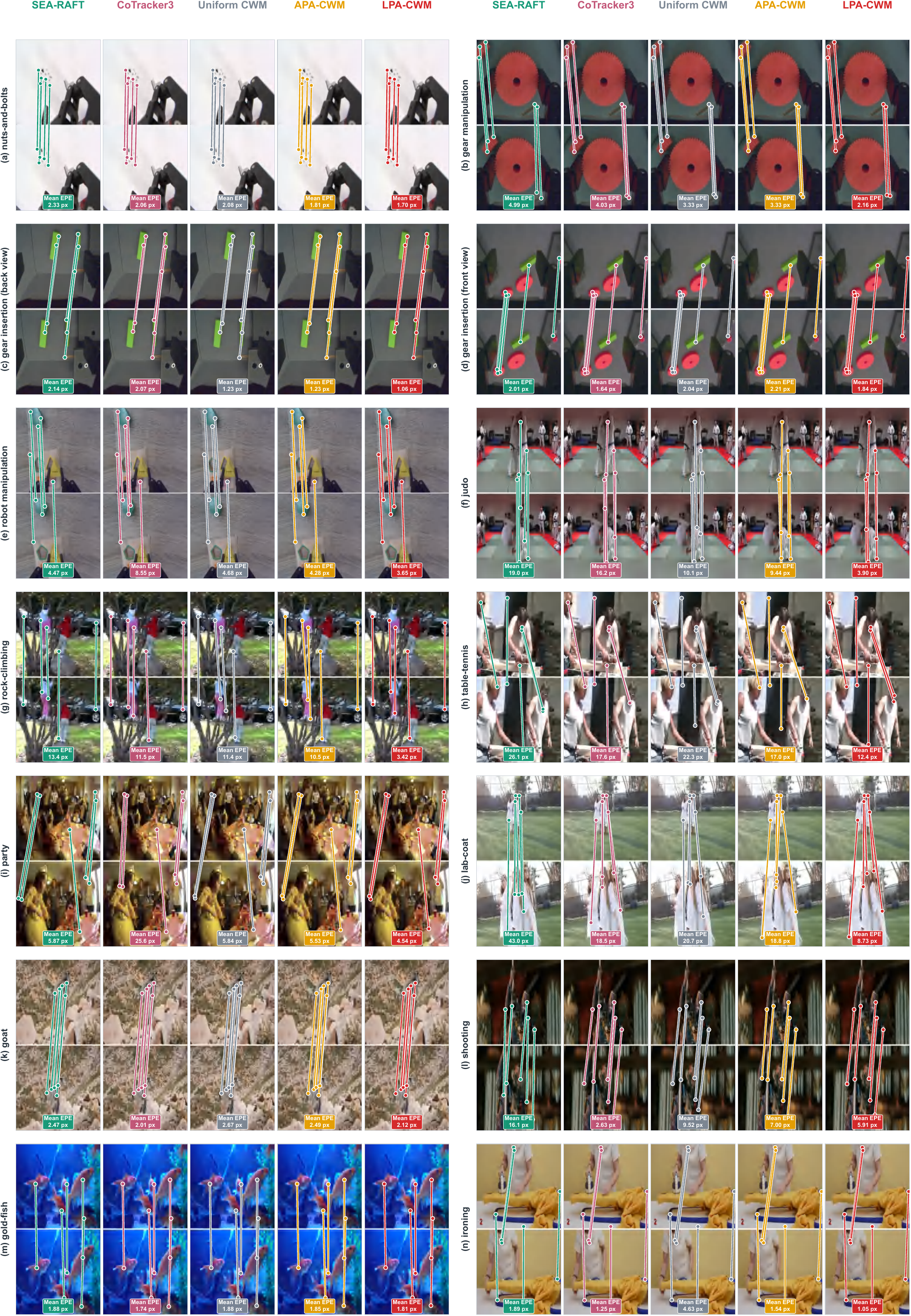}
\caption{Additional qualitative results on fourteen cases distinct from
Figure~\ref{fig:qualitative}. Columns show SEA-RAFT, CoTracker3, Uniform CWM,
APA-CWM, and LPA-CWM in this order. Every method receives the same frame pair
and the same five query identities. A method uses one color throughout;
continuous lines connect source queries to target predictions, and badges
report mean EPE over the displayed queries.}
\label{fig:additional-qualitative}
\end{figure}

\end{document}